\documentclass{article}

\usepackage[final,nonatbib]{neurips_2022}

\usepackage[utf8]{inputenc} 
\usepackage[T1]{fontenc}    
\usepackage[dvipsnames]{xcolor}
\usepackage[pagebackref=true,breaklinks=true,colorlinks,urlcolor=NavyBlue,bookmarks=false,citecolor=citecolor]{hyperref}       
\usepackage{url}            
\usepackage{booktabs}       
\usepackage{amsfonts}       
\usepackage{nicefrac}       
\usepackage{microtype}      
\usepackage{xcolor}         

\definecolor{citecolor}{HTML}{256EF6}
\definecolor{jmincolor}{HTML}{0071bc}
\definecolor{darkgreen}{HTML}{31ed09}

\usepackage{caption}
\usepackage{graphicx}
\usepackage{graphics}
\usepackage{tikz}
\usepackage{comment}
\usepackage{amsmath,amssymb} 
\usepackage{color}
\usepackage{colortbl}
\usepackage{epsfig}
\usepackage{mathtools}
\DeclareMathOperator*{\argmax}{arg\,max}

\usepackage{makecell}
\usepackage{colortbl}
\usepackage{wrapfig}
\usepackage{float}
\usepackage{bbm}
\usepackage{pifont}
\usepackage{multirow}
\usepackage{breakcites}
\usepackage{savesym}
\usepackage{wrapfig}
\savesymbol{checkmark}
\usepackage{dingbat}
\usepackage{dsfont}
\usepackage{subfig}

\def\eg{\emph{e.g.}}
\def\ie{\emph{i.e.}}
\def\etal{\emph{et al.}}

\newcommand{\cmark}{\color{darkgreen}\ding{51}\color{black}}
\newcommand{\xmark}{\color{red}\ding{55}\color{black}}
\newcommand{\cmarkb}{\ding{51}}
\newcommand{\xmarkb}{\ding{55}}
\newcommand{\norm}[1]{\left\lVert#1\right\rVert}

\newcommand\blfootnote[1]{%
  \begingroup
  \renewcommand\thefootnote{}\footnote{#1}%
  \addtocounter{footnote}{-1}%
  \endgroup
}

\definecolor{lightgray}{gray}{0.85}

\DeclareMathOperator{\EX}{\mathbb{E}}
\newcommand*{\QEDA}{\null\nobreak\hfill\ensuremath{\blacksquare}}

\title{Peripheral Vision Transformer}

\author{
    Juhong Min$^{1}$ \hspace{6mm} Yucheng Zhao$^{2,3}$ \hspace{6mm} Chong Luo$^2$ \hspace{6mm} Minsu Cho$^1$   \vspace{1.5mm}\\
    $^1$Pohang University of Science and Technology (POSTECH)\\
    $^2$Microsoft Research Asia (MSRA)\\
    $^3$University of Science and Technology of China (USTC) \vspace{2.0mm}\\
    {\small {\tt \url{http://cvlab.postech.ac.kr/research/PerViT/}}}
}

\begin{document}

\maketitle


\begin{abstract}
Human vision possesses a special type of visual processing systems called {\em peripheral vision}. 
Partitioning the entire visual field into multiple contour regions based on the distance to the center of our gaze, the peripheral vision provides us the ability to perceive various visual features at different regions.
In this work, we take a biologically inspired approach and explore to model peripheral vision in deep neural networks for visual recognition.
We propose to incorporate peripheral position encoding to the multi-head self-attention layers to let the network learn to partition the visual field into diverse peripheral regions given training data.
We evaluate the proposed network, dubbed PerViT, on ImageNet-1K and systematically investigate the inner workings of the model for machine perception, showing that the network learns to perceive visual data similarly to the way that human vision does.
The performance improvements in image classification over the baselines across different model sizes demonstrate the efficacy of the proposed method.
\end{abstract}


\section{Introduction}
\label{sec:intro}

\begin{wrapfigure}{r}{0.37\textwidth}
    \vspace{-6.0mm}
    \begin{center}
        \includegraphics[width=0.95\linewidth]{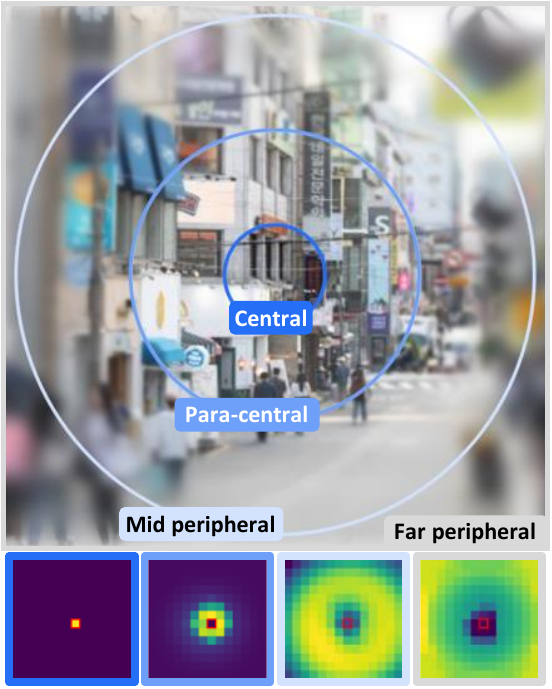}
    \end{center}
    \vspace{-2.0mm}
    \caption{This work explores blending human peripheral vision (top) with attention-based neural networks (bottom) for visual recognition.}
    \vspace{-4.0mm}
    \label{fig:teaser}
\end{wrapfigure}

For the last ten years, convolution has been a dominant feature transformation in neural networks for visual recognition due to its superiority in modelling spatial configurations of images~\cite{fukushima1982neocognitron, krizhevsky2012imagenet, lecun1998gradient}.
Despite the efficacy in learning visual patterns, the local and stationary nature of convolutional kernels limited the maximum extent of representation ability in flexible processing, \eg, dynamic transformations with global receptive fields.
Originally devised for natural language processing (NLP), self-attention~\cite{vaswani2018attention} shed a light on this direction; equipped with adaptive input processing and the ability to capture long-range interactions, it has emerged as an alternative feature transform for computer vision, being widely adopted as a core building block~\cite{dosovitskiy2021vit}.
The stand-alone self-attention models, \eg, ViT~\cite{dosovitskiy2021vit}, however, demand significantly more training data~\cite{sun2017jft} for competitive performance with its convolutional counterparts~\cite{francois2016xception, he2016deep, hu2017senet, xie2016resnext} since they miss certain desirable property which convolution possesses, \eg, locality.
These inherent pros and cons of convolution and self-attention encourage recent researches toward combinations of both so as to enjoy the best of the both worlds but which one suits the best for effective visual processing is yet controversial in literature~\cite{chu2021cpvt, cordonnier2021ontherelationship, dai2021coatnet, d2021convit, lee2022mpvit, li2022uniformer, liu2020nommer, liu2021swin, liu2022convnext, raghu2021dovitseecnn, ramachandran2019sasa, touvron2021patchconvnet, liu2022maxvit, vaswani2021halonet, wang2021pvt, wu2021cvt, xiao2021early, xu2021coat, zhao2021battle}.

Unlike the dominant visual feature transformations in machine vision, human vision possesses a special type of visual processing systems called {\em peripheral vision}~\cite{jerome1976peripheral}; it partitions the entire visual field into multiple contour regions based on the distances to the center of a gaze where each region identifies different visual aspects.
As seen in Fig.~\ref{fig:teaser}, we have high-resolution processing near the center of our gaze, \ie, central and para-central regions, to identify highly-detailed visual elements such as geometric shapes, and low-level details.
For the regions more distant from the gaze, \ie, mid and far peripheral regions, the resolution decreases to recognize abstract visual features such as motion, and high-level contexts.
This systematic strategy enables us to effectively perceive important details within a small fraction (1\%) of the visual field while minimizing unnecessary processing of background clutter in the rest (99\%), thus facilitating efficient visual processing for human brain.

According to recent study on inner workings of vision transformers~\cite{d2021convit, dosovitskiy2021vit, raghu2021dovitseecnn, touvron2021deit, xiao2021early}, their behaviors are related to the aforementioned visual processing strategies in the following respect;
the attention maps of early layers are learned to locally capture {\em fine-grained geometric details} at {\em central regions} while those of later layers perform global attentions to identify {\em coarse-grained semantics and contexts} from the whole visual field, covering {\em peripheral regions}.
This finding reveals that imitating biological designs may potentially help in modelling an effective machine vision, and also support recent approaches towards a hybrid~\cite{chen2021regionvit, dai2021coatnet, d2021convit, he2022pruning, liu2021uninet} of convolution and self-attention beyond stand-alone visual processing to take advantages of the two different perception strategies: fine-grained/local and coarse-grained/global, similarly to the human visual processing as in Fig.~\ref{fig:teaser}.

In this work, we take a biologically inspired approach and propose to inject the peripheral inductive biases\footnote{We refer peripheral inductive bias as the prioritization of our hypothesis which any attention-based neural networks can use to mimic human peripheral vision by modelling torus-shaped attentions as illustrated in Fig.~\ref{fig:teaser}.} to deep neural networks for image recognition.
We propose to incorporate peripheral attention mechanism to the multi-head self-attention~\cite{vaswani2018attention} to let the network learn to partition the visual field into diverse peripheral regions given training data where each region captures different visual features.
We experimentally show that the proposed network models effective visual periphery for reliable visual recognition.
Our main contributions can be summarized as follows:
\begin{itemize}
\vspace{-2.0mm}
\setlength\itemsep{+0.2em}
    \item This work explores to narrow the gap between human and machine vision by injecting peripheral inductive biases to self-attention layers, and presents a new form of feature transformation named \textbf{M}ulti-head \textbf{P}eripheral \textbf{A}ttention (MPA).
    \item Based on the MPA, we introduce \textbf{Per}ipheral \textbf{Vi}sion \textbf{T}ransformer (PerViT), and systematically study the inner workings of PerViT by qualitatively and quantitatively analyzing its learned attentions, which reveal that the network learns to perceive visual elements similarly to the way that human vision does without any special supervisions.
    \item The performance improvements in image classification over columnar Transformer baselines, \eg, DeiT, across different model sizes demonstrate the efficacy of the proposed approach.
\end{itemize}


\section{Related Work}
\smallbreak\noindent\textbf{Feature transformations in computer vision.}
With notable success in NLP, Transformers~\cite{devlin2019bert, vaswani2018attention} introduced a paradigm shift in computer vision~\cite{carion2020detr, caron2021dino, choromanski2021performer, dosovitskiy2021vit,  jaegle2021perceiver, ramachandran2019sasa, srinivas2021botnet, strudel2021segmenter, touvron2021deit, touvron2021patchconvnet, vaswani2021halonet}.
Despite their generalization capability, pure Vision Transformers~\cite{dosovitskiy2021vit} require extensive amount of training data to capture spatial layout of images due to lack of certain desirable property, \eg, locality.
This encouraged many recent ViT work to incorporate local inductive biases via distillation of convolutional biases~\cite{touvron2021deit}, local self-attention~\cite{lee2022mpvit, liu2021swin, xu2021coat}, a hybridization~\cite{dai2021coatnet, d2021convit, li2022uniformer}, and augmenting convolutions~\cite{chu2021cpvt, liu2022maxvit, wang2021pvt, wu2021cvt}, all of which convey a unified message:
``Despite high generalizibility of self-attentions, a sufficient amount of convolutional processing must be incorporated to capture the spatial configurations of images for reliable visual processing.''

\smallbreak\noindent\textbf{Position encoding for Transformer.}
Witnessing the efficacy of position encoding in capturing input structures in NLP~\cite{dai2019transformerXL, huang2020rpe, shaw2018selfattention}, recent vision models~\cite{dosovitskiy2021vit, ramachandran2019sasa, wu2021irpe} have begun employing position encodings for images to model spatial structures of images.
In particular, relative position encoding (RPE) plays a vital role for the purpose:
The work of Cordonnier \etal~\cite{cordonnier2021ontherelationship} proves that self-attention has close relationship with convolution when equipped with certain form of RPE.
Wu \etal~\cite{wu2021irpe} explore the existing RPE methods used in vision transformers~\cite{dai2019transformerXL, huang2020rpe, shaw2018selfattention, wang2020axialdeeplab} and draws a conclusion that RPEs impose convolutional processing on vision transformers.
Dai \etal~\cite{dai2021coatnet} observe that depthwise convolution and self-attention can naturally be unified via RPEs.
While offering promising directions, the previous RPE work, however, is limited in the sense that the focus of RPE utilization is restricted to only local attention, \eg, convolution.
This work exploits RPEs to devise an original visual feature transformation which naturally generalizes convolution and self-attention layers, thus enjoying the benefits of both via imitation of human visual processing system: 
{\em peripheral vision}.

\smallbreak\noindent\textbf{Peripheral vision for machine perception.}
Along with central vision, peripheral vision plays a vital role in a wide range of visual recognition tasks~\cite{larson2009contributions}.
The fundamental mechanisms of peripheral visual processing, however, have not been fully disclosed in human vision literature~\cite{rosenholtz2016peripheral} which stimulated many researchers to reveal its inner workings and deep implications~\cite{balas2009summary, deza2016metrics, deza2021emergent, chin2012object, rosenholtz2020demystifying, maarten2018context}.
The work of Rosenholtz~\cite{rosenholtz2016peripheral} discusses pervasive myths and current findings about peripheral vision, suggesting that peripheral vision is more crucial for human perception than previously thought to perform diverse important tasks.
Inspired by its importance, a number of pioneering work \cite{deza2021emergent, fridman2017sideeye, gould2007peripheral, harrington2022bio, lukanov2021bio, wang2017central} investigate the linkage between peripheral vision and machine vision, \eg, CNNs, while some~\cite{jonnalagadda2022foveater, vuyyuru202biologically} devise biologically-inspired models for the creation of stronger machine vision.
Continuing previous study, this paper explores to blend human peripheral vision with attention-based neural networks, \eg, vision transformer~\cite{dosovitskiy2021vit, touvron2021deit}, and introduces a new network called Peripheral Vision Transformer.



\section{Our Approach}
In this section, we introduce the Peripheral Vision Transformer (PerViT) which learns to model peripheral vision for effective image recognition. 
We first revisit the mathematical formulation of a self-attention layer and then describe how we improve it with peripheral inductive biases.


\smallbreak\noindent\textbf{Background: Multi-Head Self-Attention.}
The multi-head self-attention (MHSA)~\cite{vaswani2018attention} with $N_h$ heads performs an attention-based feature transformation by aggregating $N_h$ self-attention outputs:
\begin{align}
    \label{eq:mhsa}
    \text{MHSA}(\mathbf{X}) \coloneqq \underset{h \in [N_h]}{\text{concat}} \big[\text{Self-Attention}^{(h)}(\mathbf{X})\big] \mathbf{W}_{\text{out}} + \mathbf{b}_{\text{out}},
\end{align}
where $\mathbf{X} \in \mathbb{R}^{HW \times D_{\text{emb}}}$ is a set of input tokens and $\mathbf{W}_{\text{out}} \in \mathbb{R}^{N_{h}D_{h} \times D_{\text{emb}}}$ and $\mathbf{b}_{\text{out}} \in \mathbb{R}^{D_{\text{emb}}}$ are the transformation parameters.
The $N_h$ outputs of self-attention are designed to extract a diverse set of features from the input representation.
Formally, the self-attention at head $h$ is defined as
\begin{align}
    \label{eq:self-attention}
    \text{Self-Attention}^{(h)}(\mathbf{X}) \coloneqq \text{Normalize}\big[\Phi^{(h)}(\mathbf{X}) \big] \mathbf{V}^{(h)},
\end{align}
where $\text{Normalize}[\cdot]$ denotes a row-wise normalization and $\Phi^{(h)}(\cdot) \in \mathbb{R}^{HW \times HW}$ is a function that provides spatial attentions based on content information to aggregate the values $\mathbf{V}^{(h)} = \mathbf{X} \mathbf{W}^{(h)}_{\text{val}}$:
\begin{align}
    \label{eq:content_attention}
    \Phi^{(h)}(\mathbf{X}) \coloneqq \exp(\tau \mathbf{X} \mathbf{W}^{(h)}_{\text{qry}} (\mathbf{X} \mathbf{W}^{(h)}_{\text{key}})^{\top}) = \exp(\tau \mathbf{Q}^{(h)}, \mathbf{K}^{(h)\top}),
\end{align} 
using linear projections of $\mathbf{W}^{(h)}_{\text{qry}}, \mathbf{W}^{(h)}_{\text{key}}, \mathbf{W}^{(h)}_{\text{val}} \in \mathbb{R}^{D_{\text{emb}} \times D_{h}}$ for queries, keys, and values respectively where $\tau$ is softmax temperature and $\text{exp}(\cdot)$ applies an element-wise exponential to the input matrix.

\subsection{Multi-head Peripheral Attention}
\label{sec:mpa_main}

Based on the formulation of MHSA in Eq.~\ref{eq:mhsa}, we define \textbf{M}ulti-head \textbf{P}eripheral \textbf{A}ttention (MPA) as
\begin{align}
    \text{MPA}(\mathbf{X}) \coloneqq \underset{h \in [N_h]}{\text{concat}} \big[\text{Peripheral-Attention}^{(h)}(\mathbf{X}, \mathbf{R})\big] \mathbf{W}_{\text{out}} + \mathbf{b}_{\text{out}},
\end{align}
where $\mathbf{R} \in \mathbb{R}^{HW \times HW \times D_{\text{r}}}$ is the relative position encoding with $D_{\text{r}}$ channel dimension. The self-attention in MHSA is now replaced with Peripheral-Attention$(\cdot)$, consisting of content- and position-based attention functions $\Phi^{(h)}_{\text{c}}(\mathbf{X})$, $\Phi^{(h)}_{\text{p}}(\mathbf{R}) \in \mathbb{R}^{HW \times HW}$, which is formulated as follows:
\begin{align}
    \label{eq:peripheral-attention}
    \text{Peripheral-Attention}^{(h)}(\mathbf{X}, \mathbf{R}) \coloneqq \text{Normalize}\left[\Phi^{(h)}_{\text{c}}(\mathbf{X}) \ \odot \ \Phi^{(h)}_{\text{p}}( \mathbf{R} ) \right] \mathbf{V}^{(h)},
\end{align}
where $\odot$ is Hadamard product which mixes the given pair of attentions to provide a mixed attention 
$\Phi^{(h)}_{\text{a}}(\mathbf{X}, \mathbf{R}) \coloneqq \Phi^{(h)}_{\text{c}}(\mathbf{X}) \odot  \Phi^{(h)}_{\text{p}}(\mathbf{R}) \in \mathbb{R}^{HW \times HW}$.
For the content-based attention $\Phi_{\text{c}}$, we use exponentiated (scaled) dot-product between queries and keys as in Eq.~\ref{eq:content_attention}: $\Phi^{(h)}_{\text{c}}(\mathbf{X}) \coloneqq \exp( \tau \mathbf{Q}^{(h)} \mathbf{K}^{(h)\top})$.
For the position-based attention $\Phi_{\text{p}}$, we design a neural network that aims to imitate human visual processing system, \eg, peripheral vision, which we discuss next.

\smallbreak\noindent\textbf{Modelling peripheral vision: a Roadmap.} 
Human visual field can be grouped into several regions based on the Euclidean distances from the center of gaze, each forming ring-shaped region as seen in Fig.~\ref{fig:teaser},
where each region captures different visual aspects; the closer to the gaze, the more complex features we process, and further from the gaze, the simpler visual features we perceive.
In the context of 2-dimensional attention map $\Phi_{*}(\cdot)_{\mathbf{q},:} \in \mathbb{R}^{HW}$, we refer the query position $\mathbf{q} \in \mathbb{R}^2$ as the center of gaze, \ie, the position where feature of our interest lies at for the transformation.
We refer the local regions around the query $\mathbf{q}$ as central/para-central regions and the rest as mid/far peripheral regions.

Perhaps the simplest approach to divide the visual field into multiple subregions is to perform a single linear projection on the Euclidean distances, \ie, $\Phi^{(h)}_{\text{p}}(\mathbf{R}) = \sigma \big[ \mathbf{R}\mathbf{W}^{(h)}_{\text{p}} \big]$ where $\mathbf{W}_{\text{p}}^{(h)} \in {\mathbb{R}^{D_{\text{r}}}}$ and $\sigma[\cdot]$ is non-linearity, similarly to the previous work of Wu \etal~\cite{wu2021irpe}\footnote{Given $\sigma[\cdot] \coloneqq \exp(\cdot)$, $\text{Peripheral-Attention}^{(h)} = \text{Normalize}[ \exp( \mathbf{Q}^{(h)} \mathbf{K}^{(h)\top} ) \  \odot \ \exp(\mathbf{R}\mathbf{W}_{\text{p}}^{(h)}) ] \ \mathbf{V}^{(h)} = \text{softmax}( \mathbf{Q}^{(h)} \mathbf{K}^{(h)\top} + \mathbf{R}\mathbf{W}_{\text{rpe}}) \ \mathbf{V}^{(h)} $, which is equivalent to the {\em bias mode} RPE presented in~\cite{wu2021irpe}.}.
For straightforward imitation of peripheral vision, we use Euclidean distance for relative position input $\mathbf{R}$ and weigh the distances in $D_{\text{r}}$ different ways for the network to learn the mapping in multiple scales: $\mathbf{R}_{\mathbf{q}, \mathbf{k}, :} \coloneqq \text{concat}_{r \in [D_{\text{r}}]}[w_{\text{r}} \cdot \mathbf{R}^{\text{euc}}_{\mathbf{q}, \mathbf{k}}] \in \mathbb{R}^{D_{\text{r}}}$ where $\{w_{\text{r}}\}_{r \in [D_{\text{r}}]}$ is a set of learnable parameters shared across layers and heads, and $\mathbf{R}^{\text{euc}}_{\mathbf{q}, \mathbf{k}} = \norm{\mathbf{q} - \mathbf{k}}_{2}$ is the Euclidean distance between query and key positions $\mathbf{q}, \mathbf{k} \in \mathbb{R}^{2}$.
In our experiments, we choose sigmoid function for $\sigma$ to provide normalized (positive) weights to the content-based attention $\Phi_{\text{c}}$.
A main drawback of this single-layer formulation is that $\Phi_{\text{p}}$ is only able to provide Gaussian-like attention map as seen in top-left in Fig.~\ref{fig:peripheral_rep}, thus being unable to represent peripheral regions in diverse shapes.
For the encoding function to represent various (torus-shaped) peripheral regions, the distances must be processed by an MLP:
\begin{align}
    \label{eq:mlp_phi_p}
    \Phi^{(h)}_{\text{p}}(\mathbf{R}) = \sigma \left[ \text{Linear} ( \text{ReLU} ( \text{Linear} (\mathbf{R}; \mathbf{W}_{\text{p1}} )) \mathbf{W}^{(h)}_{\text{p2}}) \right] = \sigma \left[ \text{ReLU} ( \mathbf{R}\mathbf{W}_{\text{p1}} ) \mathbf{W}^{(h)}_{\text{p2}} \right],
\end{align}
where $\mathbf{W}_{\text{p1}} \in \mathbb{R}^{D_{\text{r}} \times D_{\text{hid}}}$ and $\mathbf{W}^{(h)}_{\text{p2}} \in \mathbb{R}^{D_{\text{hid}}}$ are the linear projection parameters\footnote{We omit the bias terms in the linear layers for brevity.}, and ReLU gives non-linearity to the function.
The first projection $\mathbf{W}_{\text{p1}}$ is shared across the heads in order to exchange information so each function is able to provide attention that are effective or complementary to other heads' attention.
Note that given identical relative distances between a fixed query point $\mathbf{q} \in \mathbb{R}^{2}$ and key points $\mathbf{k}_{i}, \mathbf{k}_{j} \in \mathbb{R}^2$, \ie, $\mathbf{R}_{\mathbf{q}, \mathbf{k}_{i}} = \mathbf{R}_{\mathbf{q}, \mathbf{k}_{j}}$, Eq.~\ref{eq:mlp_phi_p} provides the same attention scores: $\Phi_{\text{p}}(\mathbf{R})_{\mathbf{q}, \mathbf{k}_i} = \Phi_{\text{p}}(\mathbf{R})_{\mathbf{q}, \mathbf{k}_j}$ as seen in top-right of Fig.~\ref{fig:peripheral_rep}.
This property, however, is not always desired in practical scenarios because the rotational symmetric property hardly holds for most real-world objects.
To break the symmetric property in Eq.~\ref{eq:mlp_phi_p} while preserving peripheral design to sufficient extent, we introduce \textbf{{\em peripheral projections}} in which the transformation parameters are given small spatial resolutions, \eg, $K \times K$ window, such that $\mathbf{W}_{\text{p1}} \in \mathbb{R}^{K^2 \times D_{\text{r}} \times D_{\text{hid}}}$ and $\mathbf{W}^{(h)}_{\text{p2}} \in \mathbb{R}^{K^2 \times D_{\text{hid}}}$ so that they provide similar but different attention scores, $\Phi_{\text{p}}(\mathbf{R})_{\mathbf{q}, \mathbf{k}_i} \neq \Phi_{\text{p}}(\mathbf{R})_{\mathbf{q}, \mathbf{k}_j}$, given $\mathbf{R}_{\mathbf{q}, \mathbf{k}_{i}} = \mathbf{R}_{\mathbf{q}, \mathbf{k}_{j}}$, by aggregating neighboring relative distances around the key location $\mathbf{k}$ as follows:
\begin{align}
    \label{eq:phi_p}
    \Phi^{(h)}_{\text{p}} (\mathbf{R})_{\mathbf{q}, \mathbf{k}, :} \coloneqq \sigma \left[ \sum_{\mathbf{n} \in \mathcal{N}(\mathbf{k})}\text{ReLU} \left( \sum_{\mathbf{m} \in \mathcal{N}(\mathbf{k})} \mathbf{R}_{\mathbf{q}, \mathbf{m}, :} \mathbf{W}_{\text{p1} \ \mathbf{m}-\mathbf{k}, :, :} \right) \mathbf{W}^{(h)}_{\text{p2} \ \mathbf{n}-\mathbf{k}, :} \right],
\end{align}
where $\mathcal{N}(\mathbf{k}) \coloneqq \left[ \mathbf{k}-\lfloor\frac{K}{2}\rfloor, \dots, \mathbf{k}+\lfloor\frac{K}{2}\rfloor \right] \times \left[ \mathbf{k}-\lfloor\frac{K}{2}\rfloor, \dots, \mathbf{k}+\lfloor\frac{K}{2}\rfloor \right]$ is a set of $K^2$ neighbors around input position $\mathbf{k}$. We set $K=3$ for all layers and heads as $K>3$ hardly brought improvements.
Note that each linear projection in Eq.~\ref{eq:phi_p} is equivalent to a 4-dimensional convolution~\cite{rocco2018neighbourhood}, taking 4-dimensional input $\mathbf{R} \in \mathbb{R}^{HW \times HW \times D_{\text{r}}}$ to process in convolutional manner using 4-dimensional kernels in size of $K \times K \times 1 \times 1$, \ie, $\mathbf{W}_{\text{p1}} \in \mathbb{R}^{K \times K \times 1 \times 1 \times D_{\text{r}} \times D_{\text{hid}}}$.
Precisely, the peripheral projection considers a small subset of 4D local neighbors that pivots the query position $\mathbf{q}$, similarly to the center-pivot 4D convolution~\cite{min2021chm, min2021hypercorrelation}.
After each peripheral projection, we add an instance normalization layer~\cite{ulyanov2017instancenorm} for stable optimization:
\begin{align}
    \mathbf{R}' = \text{ReLU} \left( \text{IN} ( \text{PP} ( \mathbf{R}; \mathbf{W}_{\text{p1}} ); \pmb{\gamma}_{\text{p1}}, \pmb{\beta}_{\text{p1}} )\right), \ \ \ \ \ \Phi^{(h)}_{\text{p}} (\mathbf{R}) = \sigma \left( \text{IN} ( \text{PP} ( \mathbf{R}'; \mathbf{W}_{\text{p2}}^{(h)} ); {\gamma}^{(h)}_{\text{p2}}, {\beta}^{(h)}_{\text{p2}} ) \right),
\end{align}
where $\pmb{\gamma}_{\text{p1}}, \pmb{\beta}_{\text{p1}} \in \mathbb{R}^{D_{\text{hid}}}$ and $\gamma^{(h)}_{\text{p2}}, \beta^{(h)}_{\text{p2}} \in \mathbb{R}$ are weights/biases of the instance norms and $\text{PP}(\cdot)$ denotes the peripheral projection: $\text{PP}(\mathbf{R}, \mathbf{W})_{\mathbf{q}, \mathbf{k}, :} \coloneqq \sum_{\mathbf{n} \in \mathcal{N}(\mathbf{k})} \mathbf{R}_{\mathbf{q}, \mathbf{n}, :} \mathbf{W}_{\mathbf{n} - \mathbf{k}, :, :}$.
The middle row of Fig.~\ref{fig:peripheral_rep} depicts learned attentions of $\Phi_{\text{p}}$ with peripheral projections, which provides peripheral attention maps in greater diversity compared to single- and multi-layer counterparts without $\mathcal{N}$.

\begin{table}[tp] 
	\begin{minipage}{0.47\linewidth}
		\centering
        \scalebox{0.85}{
            \includegraphics[width=1.15\textwidth]{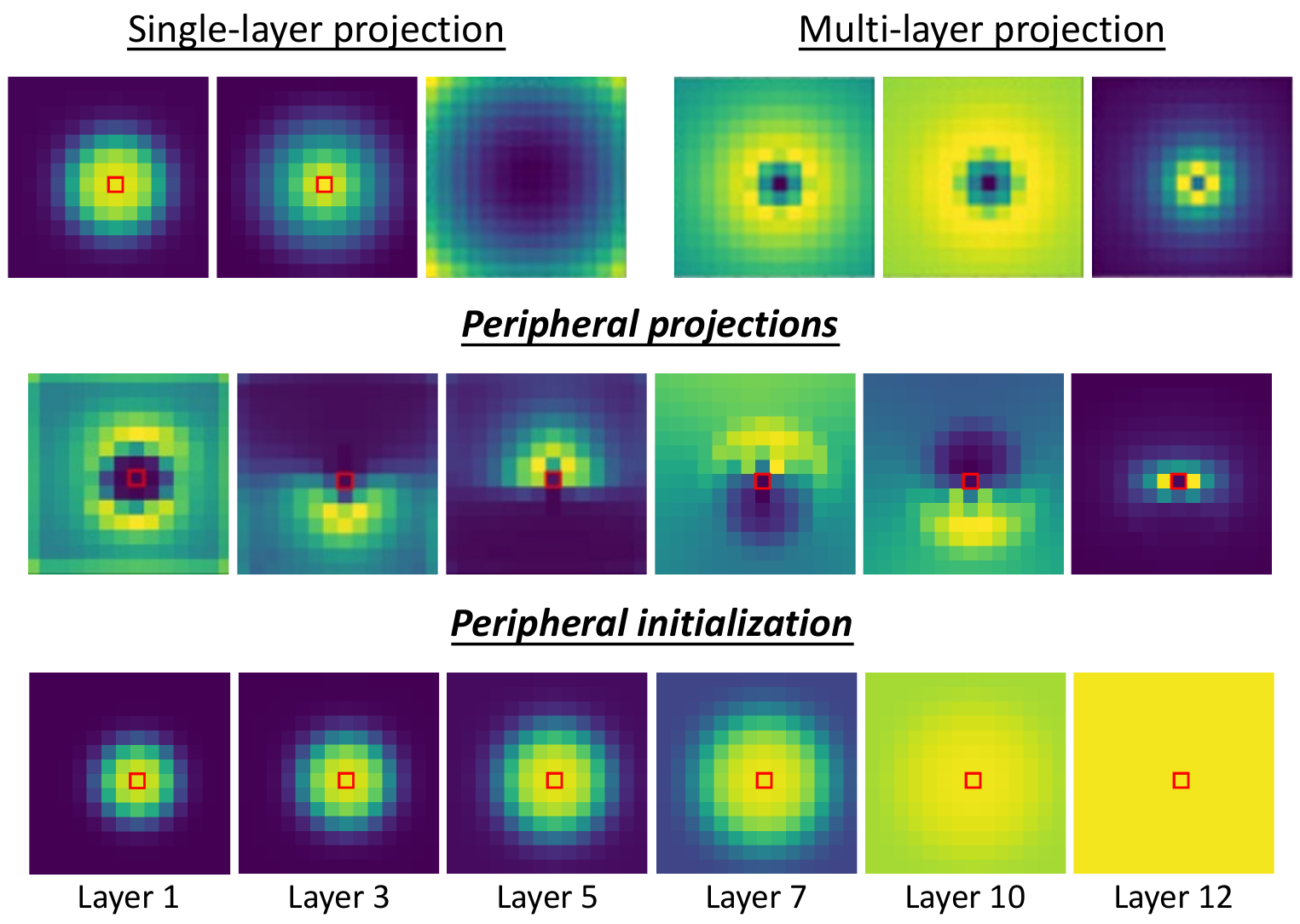}
        }
        \vspace{1.0mm}
		\captionof{figure}{Top: representation ability of $\Phi_{\text{p}}$ under varying \# layers. Middle: peripheral projections. Bottom: peripheral initialization.}
		\label{fig:peripheral_rep}
	\end{minipage}\hfill
	\begin{minipage}{0.5\linewidth}
		\centering
        \scalebox{0.95}{
            \includegraphics[width=1.05\textwidth]{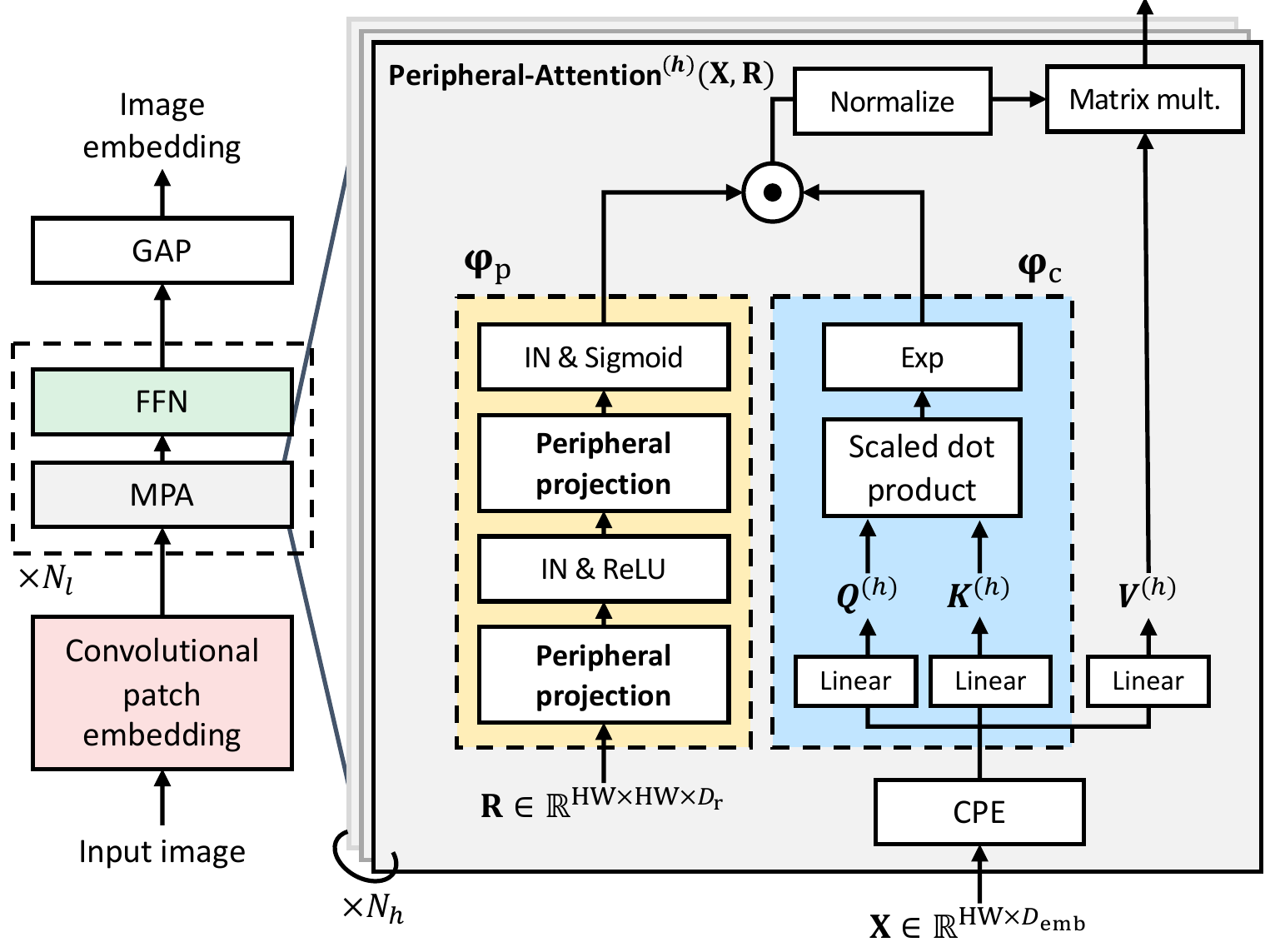}
		}
		\vspace{-2.0mm}
		\captionof{figure}{Overall architecture of PerViT which is based on DeiT~\cite{d2021convit} architecture.}
        \label{fig:architecture}
	
	\end{minipage}
	\vspace{-4.0mm}
\end{table}

\smallbreak\noindent\textbf{Peripheral initialization.}
Recent study~\cite{raghu2021dovitseecnn, touvron2021deit} observe that early layers of trained vision transformer learn to attend locally whereas late layers perform global attentions.
To facilitate training of our network, we inject this property in the beginning of the training stage by initializing parameters of $\Phi_{\text{p}}$ for the purpose, making attention scores near the queries larger than the distant ones in the early layers while uniformly distributing them in the late layers as seen in bottom row of Fig.~\ref{fig:peripheral_rep}.
We refer this method as \textbf{{\em peripheral initialization}} for its resemblance to the functioning of peripheral vision~\cite{jerome1976peripheral} which also operates either locally or globally to perceive different visual features~\cite{zeiler2014visual}.
To formally put, given two arbitrarily chosen distances $\mathbf{R}^{\text{euc}}_{\mathbf{q}, \mathbf{k}_i}, \mathbf{R}^{\text{euc}}_{\mathbf{q}, \mathbf{k}_j} \in \mathbb{R}$ which satisfy $\mathbf{R}^{\text{euc}}_{\mathbf{q}, \mathbf{k}_i} < \mathbf{R}^{\text{euc}}_{\mathbf{q}, \mathbf{k}_j}$, we want $\Phi^{(l, h)}_{\text{p}}(\mathbf{R})_{\mathbf{q}, \mathbf{k}_i} \gg \Phi^{(l, h)}_{\text{p}}(\mathbf{R})_{\mathbf{q}, \mathbf{k}_j}$\footnote{We now use the superscript to indicate both layer and head indices for the ease of demonstration.} as $l \rightarrow 1$, \ie, local attention in early layers, and $\Phi^{(l,h)}_{\text{p}}(\mathbf{R})_{\mathbf{q}, \mathbf{k}_i} \approx \Phi^{(l,h)}_{\text{p}}(\mathbf{R})_{\mathbf{q}, \mathbf{k}_j}$ as $l \rightarrow N_l$, \ie, global attention in late layers, where $N_l$ is the total number of MPA layers.
We first initialize the parameters of $\Phi^{(l,h)}_{\text{p}}$ and $\{w_r\}_{r \in [D_r]}$ to particular values.
Specifically, for all layers $l \in [N_l]$ and heads $h \in [N_h]$,
\begin{align}
    w_r \coloneqq -c_{1},  \forall r \in [D_{\text{r}}] \ \ \ \ \ \mathbf{W}^{(l)}_{\text{p1}} \coloneqq c_{2}J_{K^{2}, D_{\text{r}}, D_{\text{hid}}} \ \ \ \ \ \mathbf{W}^{(l,h)}_{\text{p2}} \coloneqq c_{2}J_{K^{2}, D_{\text{hid}}} \ \ \ \ \ \pmb{\gamma}^{(l)}_{\text{p1}} \coloneqq \mathbf{1} \ \ \ \ \ \pmb{\beta}^{(l)}_{\text{p1}} \coloneqq \mathbf{0}
\end{align}
where $c_1, c_2 \in \mathbb{R}^{+}$ are positive reals, and $J_{N, M} \in \mathbb{R}^{N \times M}$ refers to all-one matrix in size $N \times M$.
The above initialization provides local attention after the second peripheral projection, \ie, $\text{PP}(\mathbf{R}'; \mathbf{W}_{\text{p2}}^{(h)} )_{\mathbf{q}, \mathbf{k}_{i}, :} > \text{PP}(\mathbf{R}'; \mathbf{W}_{\text{p2}}^{(h)})_{\mathbf{q}, \mathbf{k}_{j}, :}$ given $\mathbf{R}^{\text{euc}}_{\mathbf{q}, \mathbf{k}_i} < \mathbf{R}^{\text{euc}}_{\mathbf{q}, \mathbf{k}_j}$.
Next, based on our findings that biases $\beta^{(l,h)}_{\text{p2}}$ and the weights $\gamma^{(l,h)}_{\text{p2}}$ in the second instance norm control the sizes and strengths of local attention respectively, we simulate peripheral initialization by setting their initial values as $\beta^{(l,h)}_{\text{p2}} \coloneqq s_l$ and $\gamma^{(l,h)}_{\text{p2}}  \coloneqq v_l$ where respective $\{s_l\}_{l \in [N_l]}$ and $\{v_l\}_{l \in [N_l]}$ are sets of initial values for attention sizes and strengths.
We set their values collected from uniform intervals: $s_l \in [-5.0, 4.0]$ and $v_l \in [3.0, 0.01]$ where $s_{l-1} < s_l$ and $v_{l-1} > v_l$ to give  stronger local attentions to shallow layers compared to deep ones as seen in bottom row of Fig.~\ref{fig:peripheral_rep}.
We set $c_{1}, c_{2}=0.02$ in our experiments.
We refer the readers to the Appendix~\ref{sec:supp_peripheral_init} for the complete derivation of the peripheral initialization.

\subsection{Peripheral Vision Transformer}
Based on the proposed peripheral projections and initialization, we develop image classification models, dubbed Peripheral Vision Transformer, which is illustrated in Fig.~\ref{fig:architecture}.
We follow similar architecture to DeiT~\cite{touvron2021deit} with convolutional patch embedding stem;
as the original patchify stem~\cite{dosovitskiy2021vit} exhibits substandard optimizability due to its coarse-grained early visual processing~\cite{xiao2021early}, many recent ViT models adopt multi-resolution {\em pyramidal designs}~\cite{liu2021swin, wang2021pvt, wu2021cvt, xu2021coat} to mitigate the issue.
While the pyramidal models have shown their efficacy in learning reliable image embeddings, we stick with the original single-resolution {\em columnar design} for PerViT because features in multiple resolution make our study less interpretable, which further requires additional techniques for combining our peripheral attention $\Phi_{\text{p}}$ with the existing cost-effective self-attentions mechanisms such as factorized attention~\cite{xu2021coat}, shifted-window~\cite{liu2021swin}, and cross-shaped window~\cite{dong2021cswin}.
To carry out fine-grained early processing while keeping single-resolution features across the layers, we adopt convolutional patch embedding layer~\cite{xiao2021early} with multi-stage layouts for channel dimensions.
The convolutional embedding layer consists of four $3 \times 3$ and one $1 \times 1$ convolutions where the $3 \times 3$ convolutions are followed by batch norm~\cite{ioffe2015batchnorm} and ReLU~\cite{nair2010relu}.
For additional architecture details, we refer to the Appendix~\ref{sec:supp_layout_details}.

\smallbreak\noindent\textbf{Overall pipeline.}
Given an image, the convolutional patch embedding provides token embeddings $\mathbf{X}^{(1)} \in \mathbb{R}^{HW \times D_{\text{emb}}}$.
Similarly to~\cite{dosovitskiy2021vit, touvron2021deit}, the embeddings are fed to a series of $N_l$ blocks each of which consists of an MPA layer and a feed-forward network with residual pathways:
\begin{align}
    \mathbf{X}^{(l')} = \text{MPA}(\text{LN}(\text{CPE}(\mathbf{X}^{(l)}))) + \mathbf{X}^{(l)}, \ \ \ \ \ \ \ \ \ 
    \mathbf{X}^{(l+1)} = \text{FFN}(\text{LN}(\mathbf{X}^{(l')})) + \mathbf{X}^{(l')},
\end{align}
where $\text{LN}$ is layer normalization~\cite{ba2016layernorm}, and $\text{FFN}$ is an MLP consisting of two linear transformations with a GELU activation~\cite{hendrycks2020gelu}.
Following the work of~\cite{lee2022mpvit, xu2021coat}, we adopt convolutional position encoding (CPE), \ie, a $3 \times 3$ depth-wise convolution, before first layer norm for its efficacy with negligible computational cost.
The output $\mathbf{X}^{(N_h)}$ is global-average pooled to form an image embedding.


\section{Experiments}
In this section, we first investigate the inner workings of PerViT trained on ImageNet-1K classification dataset to examine how it benefits from the proposed peripheral projections and initialization, and then compare the method with previous state of the arts under comparable settings.

\smallbreak\noindent\textbf{Experimental setup.} 
Our experiments focus on image classification on ImageNet-1K~\cite{deng2009imagenet}.
Following training recipes of DeiT~\cite{touvron2021deit}, we train our model on ImageNet-1K from scratch with batch size of 1024, learning rate of 0.001 using AdamW~\cite{loshchilov2019adamw} optimizer, cosine learning rate decay scheduler, and the same data augmentations~\cite{devlin2019bert} for 300 epochs, including warm-up epochs.
We evaluate our model with three different sizes, \eg, Tiny (T), Small (S), and Medium (M).
We use stochastic depths of 0.0, 0.1, and 0.2 for T, S, and M respectively.
We refer to the Appendix~\ref{sec:supp_training_details} for additional details. 

\begin{table}[tp] 
	\begin{minipage}{0.57\linewidth}
		\centering
        \scalebox{0.99}{
            \includegraphics[width=0.99\textwidth]{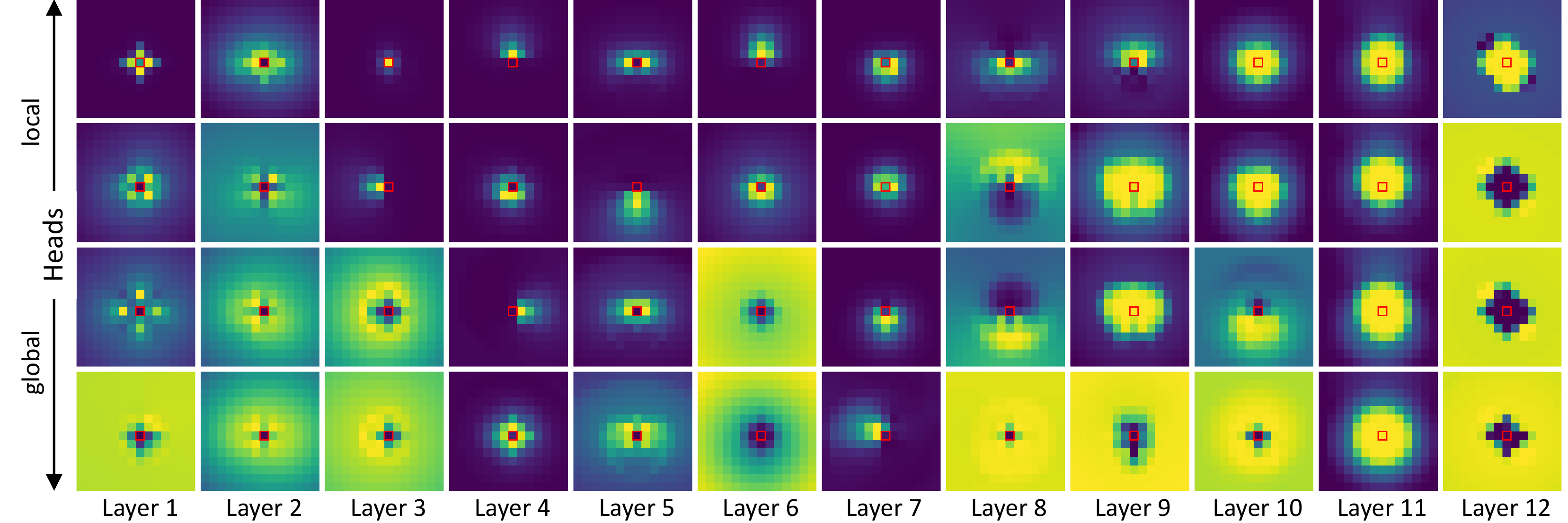}
        }
        \vspace{0.5mm}
		\captionof{figure}{Learned attentions $\Phi_{\text{p}}$ of PerViT-T.}
		\label{fig:tiny_weight}
	\end{minipage}\hfill
	\begin{minipage}{0.41\linewidth}
		\centering
        \scalebox{0.99}{
        \includegraphics[width=0.99\textwidth]{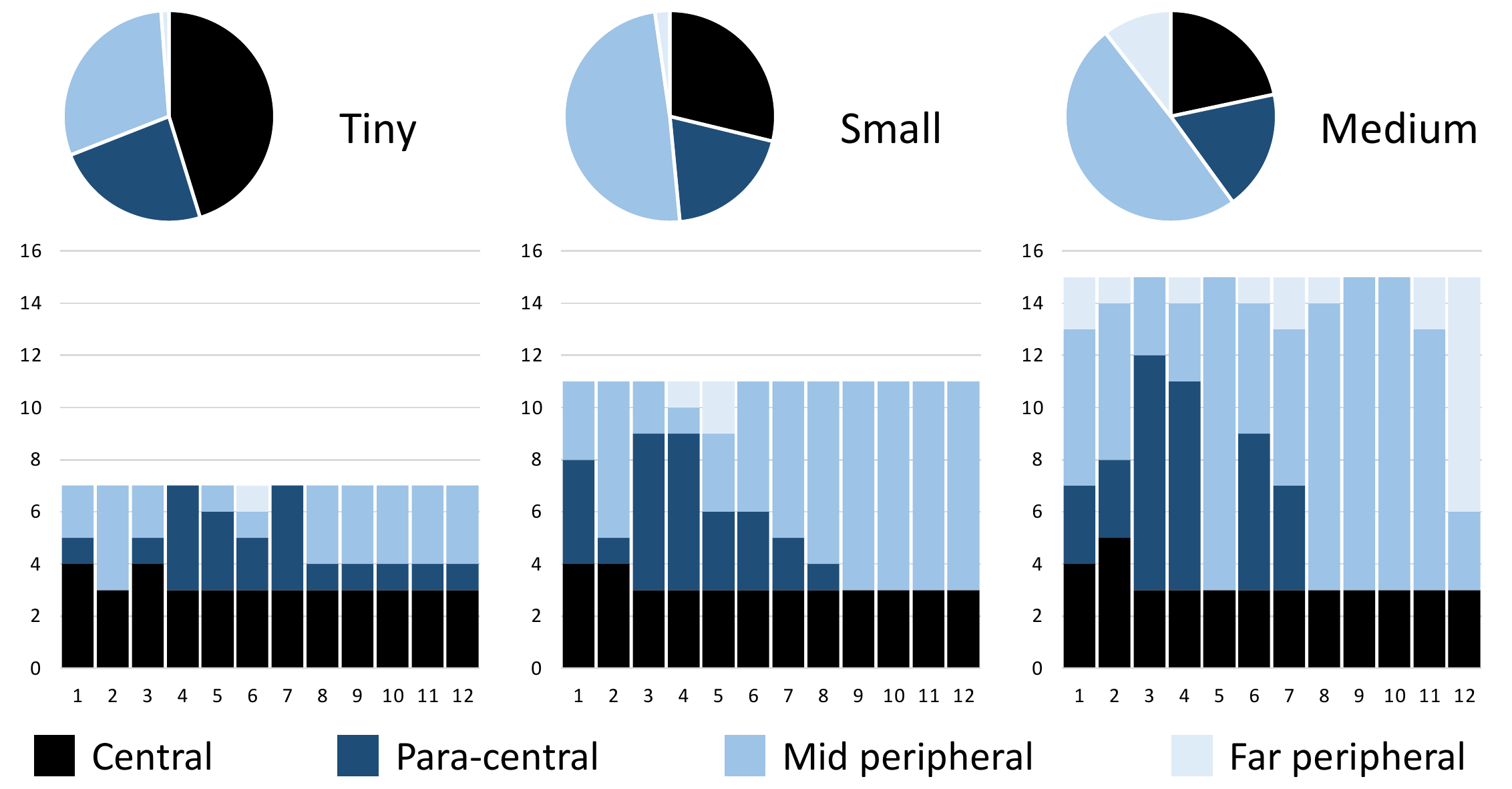}
		}
		\vspace{-2.0mm}
		\captionof{figure}{Peripheral region classification.}
        \label{fig:peri_cls}
	
	\end{minipage}
	\vspace{-3.0mm}
\end{table}

\begin{figure*}[t]
    \begin{center}

    \scalebox{0.33}{
    \centering
        \includegraphics{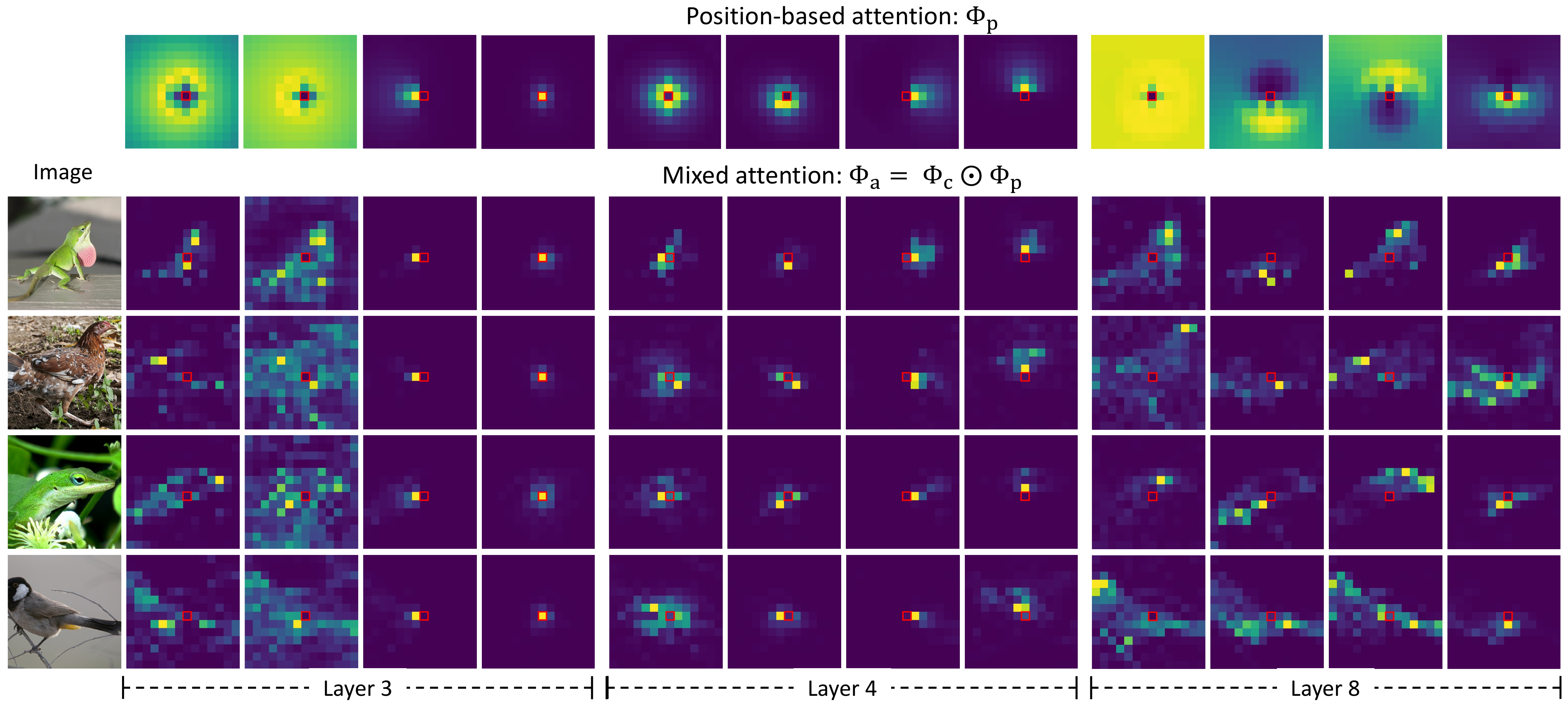}
    }
    \caption{Visualization of learned attentions $\Phi_{\text{p}}$ and $\Phi_{\text{a}}$ for $l \in \{3, 4, 8\}$. Best viewed in electronics.} 
    \vspace{-3.0mm}
    \label{figure:qualitative_study}
    \end{center}   
\end{figure*}

\subsection{The inner workings of PerViT}
\label{sec:innerworking_main}
\smallbreak\noindent\textbf{Learning peripheral vision.} 
We begin by investigating how PerViT models peripheral vision by qualitatively analyzing its learned attention of $\Phi_{\text{p}}$.
Figure~\ref{fig:tiny_weight} depicts the learned attention map of $\Phi^{(l, h)}_{\text{p} \ \mathbf{q}, :} \in \mathbb{R}^{HW}$ for all layers and heads where the query position is given at the center, \ie, $\mathbf{q} = [7, 7]^{\top}$\footnote{The columnar design of PerViT provides identical spatial resolution for every intermediate feature map in the network: $H, W = 14$, thus facilitating the ease of qualitative/quantitative analyses of the learned attentions.}.
We observe that the attentions are learned to be in diverse shapes of peripheral regions.
Interestingly, without any special supervisions, the four attended regions ($N_h = 4$) in most layers are learned to complement each other to cover the entire visual field, capturing different visual aspects at each region (head), similarly to human peripheral vision illustrated in Fig.~\ref{fig:teaser}.
For example, first two heads in Layer 3 attend the central regions while the others cover the rest peripheral regions.
The second and third heads in Layer 8 cover top and bottom hemicircles respectively, forming a circular-shaped semi-global receptive field.
Moreover, a large number of early attentions is in form of central/para-central regions while those of late layers are learned to cover mid to far peripheral regions.
To quantitatively inspect how PerViT models the peripheral visual system, we classify every feature transformation layer in the network into one of the four visual regions $\mathbb{P} \in \{\text{c}, \text{p}, \text{m}, \text{f}\}$ where respective elements refer to central, para-central, mid, and far peripheral regions.
PerViT-Attention of head $h$ at layer $l$ is classified as peripheral region $p$ if the average of its attention scores which fall in visual region $p$ is the largest among the others: 
\begin{align}
    \text{PeripheralRegion}{(l,h)} \coloneqq \argmax_{p \in \mathbb{P}}\left[ \frac{1}{|\mathcal{P}|^{2}} \sum_{(\mathbf{q}, \mathbf{k}) \in \mathcal{P} \times \mathcal{P}} \Phi^{(l,h)}_{\text{p} \ \ \mathbf{q}, \mathbf{k}} \cdot \mathbbm{1}\left[\norm{\mathbf{q} - \mathbf{k}}_{2} \in \mathcal{I}_{p}\right] \right],
\end{align}
where $\mathcal{P}$ is a set of spatial positions ($|\mathcal{P}| = HW$) and $\mathcal{I}_{p}$ is distance range (real-valued interval) of peripheral region $p$\footnote{We use $\mathcal{I}_{\text{c}} = [0, 1.19)$ , $\mathcal{I}_{\text{p}} = [1.19, 3.37)$, $\mathcal{I}_{\text{m}} = [3.37, 5.83)$, and $\mathcal{I}_{\text{f}} = [5.83, 7.9)$. We refer the readers to the Appendix~\ref{sec:supp_peripheral_region} for the justification on the these interval choices.}.
The pie charts of Fig.~\ref{fig:peri_cls} describe the proportions of peripheral regions for Tiny, Small, and Medium models where the bar graphs show them in layer-wise manner\footnote{We classify the $3 \times 3$ depth-wise convolution in CPE and the two linear projections in FFN as central regions as their receptive fields approximately fall in the interval of $\mathcal{I}_{\text{c}} = [0, 1.19)$.}.
Similarly to the visualized attention maps in Fig.~\ref{fig:tiny_weight}, the early layers attends central/para-central regions whereas deeper ones focus on outer region.
We observe that, as the model size grows, the number of mid/far peripheral attention increases whereas that of central/para-central attention stays similar, suggesting that the models no longer require local attentions once sufficient amount of processing is done in the central region because, we hypothesize, identifying geometric patterns, \eg, corners and edges, is relatively simpler process than understanding high-level semantics.

\begin{figure*}[t]
    \begin{center}

    \scalebox{0.34}{
    \centering
    \includegraphics{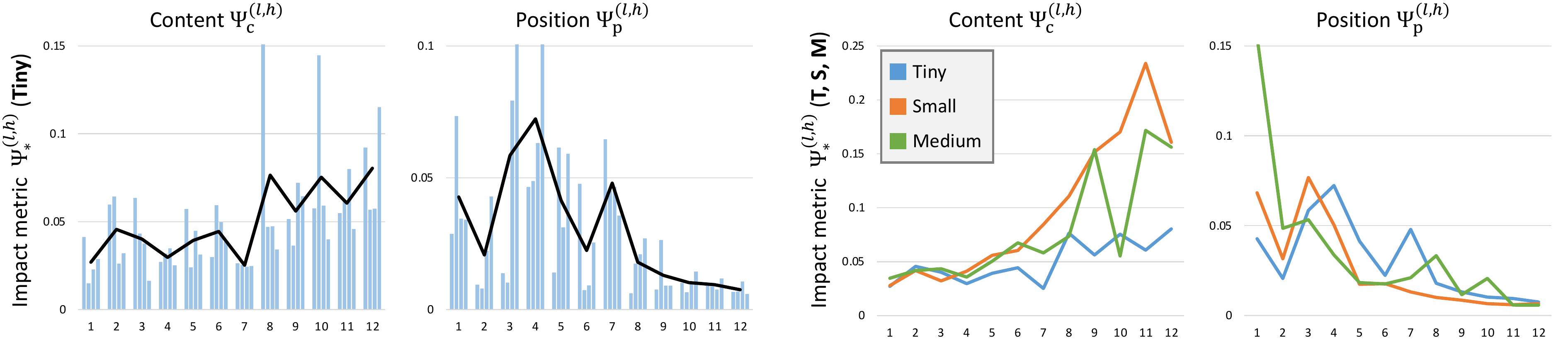}
    }
    \caption{The measure of impact (x-axis: layer index, y-axis: the impact metric $\Psi_{*}$). Each bar graph shows the measure of a single head (4 heads at each layer), and the solid lines represent the trendlines which follow the average values of layers. (left: results of PerViT-T. right: results of T, S, and M.)}
    \label{fig:phiratio}
    \end{center}    
\end{figure*}

\smallbreak\noindent\textbf{Inspecting the impact of attentions (static {\em vs.} dynamic).}
To study how position-based attentions $\Phi_{\text{p}}$ contribute to the mixed attentions $\Phi_{\text{a}} = \Phi_{\text{c}} \odot \Phi_{\text{p}}$, we collect sample images and visualize their attention maps of Layers 3, 4 and 8 in Fig.~\ref{figure:qualitative_study}.
The mixed attentions $\Phi_{\text{a}}$ at Layer 4 are formed dynamically ($\Phi_{\text{c}}$) within statically-formed region ($\Phi_{\text{p}}$) while the attentions $\Phi_{\text{a}}$ at Layer 8 weakly exploit position information ($\Phi_{\text{p}}$) to form dynamic attentions ($\Phi_{\text{c}}$).
The results reveal that $\Phi_{\text{p}}$ plays two different roles; it imposes \textbf{{\em semi-dynamic attention}} if the attended region is focused in a small area whereas it serves as \textbf{{\em position bias injection}} when the attended region is relatively broad.
In the Appendix~\ref{sec:supp_mpa_conv_mhsa}, we constructively prove that {\em an MPA layer in extreme case of semi-dynamic attention/position bias injection is in turn convolution/multi-head self-attention}, naturally generalizing the both transformations.
To quantitatively examine the contributions of $\Phi_{\text{c}}$ and $\Phi_{\text{p}}$ to the mixed attention $\Phi_{\text{a}}$, we define a measure of `impact' by taking inverse of difference between two attentions:
\begin{align}
    \Psi_{\text{p}}^{(l,h)} \coloneqq  [ ||\Phi_{\text{a}}^{(l, h)} - \Phi_{\text{p}}^{(l, h)}||_{F} ]^{-1},
\end{align}
where $\norm{\cdot}_{F}$ is Frobenius norm.
The higher the measure $\Psi_{\text{p}}^{(l,h)}$, the larger the impact of position-based attention $\Phi_{\text{p}}^{(l,h)}$.
Being averaged over all test samples, $\Psi_{\text{c}}^{(l,h)}$ is similarly defined.
As seen in Fig.~\ref{fig:phiratio}, we observe a clear tendency that the impact of position-based attention is significantly higher in early processing, transforming features {\em semi-dynamically}, while the later layers require less position information, regarding $\Phi_{\text{p}}$ as a minor {\em position bias}.
This tendency becomes more visible with larger models as seen in right of Fig~\ref{fig:phiratio}; Small and Medium models exploit dynamic transformations much more compared to Tiny model, especially in the later layers.
Moreover, we note that the impact measures of four heads (bar graphs) within each layer are unevenly distributed, showing high variance, which imply that the network evenly utilizes both position and content information simultaneously within each MPA layer as seen in Layer 3 in Fig.~\ref{figure:qualitative_study}, {\em performing both static/local and dynamic/global transformation in a single-shot}.
These results reveal that feature transformations for effective visual recognition should not be restricted to be in either position-only~\cite{he2016deep, liu2022convnext} or content-only~\cite{dosovitskiy2021vit, touvron2021deit} design but they should be in the form of a hybrid~\cite{dai2021coatnet, d2021convit}.

\begin{figure*}[t]
    \begin{center}

    \scalebox{0.34}{
    \centering
    \includegraphics{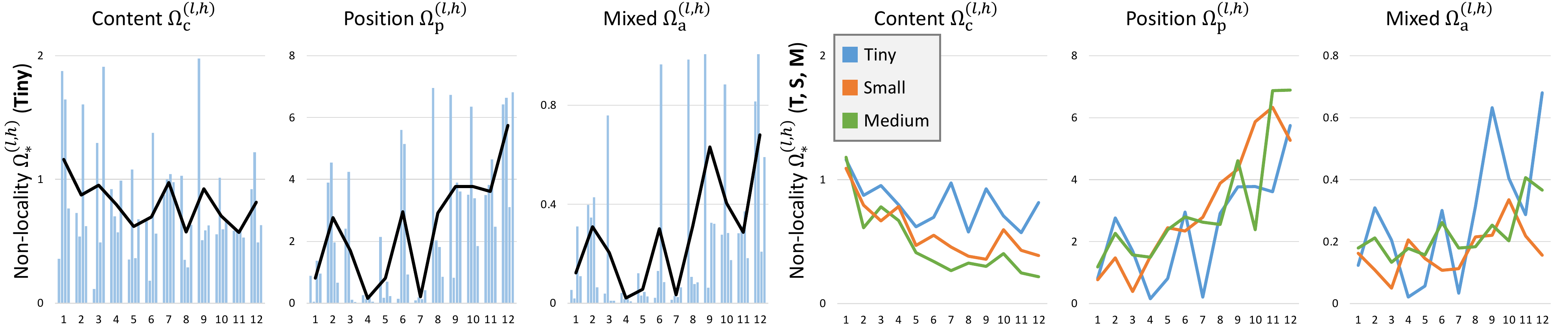}
    }
    \caption{The measure of nonlocality (x-axis: layer index, y-axis: the nonlocality metric $\Omega_{*}$).}
    \label{fig:nonlocality} 
    \end{center}    
    \vspace{-3.0mm}
\end{figure*}

\smallbreak\noindent\textbf{Inspecting the locality (local {\em vs.} global).}
We further investigate the inner workings of PerViT by quantifying how locally $\Phi_{*}$ attends.
Following the work of~\cite{d2021convit}, we define the measure of nonlocality for $\Phi^{(l,h)}_{\text{p}}$ by summing all pair-wise query-key distances weighted by their attention scores: 
\begin{align}
    \label{eq:nonlocality}
    \Omega_{\text{p}}^{(l,h)} \coloneqq \frac{1}{|\mathcal{P}|^{2}}\sum_{(\mathbf{q},\mathbf{k}) \in \mathcal{P} \times \mathcal{P}} \Phi_{\text{p} \ \mathbf{q},\mathbf{k}}^{(l, h)} \mathbf{R}_{\mathbf{q},\mathbf{k}}^{\text{euc}}.
\end{align}
The metrics $\Omega_{\text{c}}$ and $\Omega_{\text{a}}$ are similarly defined, being averaged over all test samples.
As seen in Fig.~\ref{fig:nonlocality}, we observe a similar trend of locality between $\Phi_{\text{p}}$ and $\Phi_{\text{a}}$, which reveals the position information play more dominant role over the content information in forming spatial attentions ($\Phi_{\text{a}}$) for feature transformation.
Interestingly, we also observe that content- and position-based attentions behave conversely;
$\Phi_{\text{c}}$ attends globally in early layers, \ie, large scores are distributed over the whole spatial region, while being relatively local in deeper layers.
We hypothesize that the proposed $\Phi_{\text{p}}$ in early layers is trained to effectively suppress unnecessary scores of $\Phi_{\text{c}}$ at distant positions, thus exploiting only a few relevant ones within its local region of interest.
Meanwhile, $\Phi_{\text{p}}$ at later layers gives $\Phi_{\text{c}}$ higher freedom in forming the spatial attention $\Phi_{\text{a}}$ as described in the plots of Fig.~\ref{fig:phiratio}, which allows the attention scores of $\Phi_{\text{c}}$ to be clustered in semantically meaningful parts, \eg, eyes of the animals as seen in attentions of the first head at Layer 8 (Fig.~\ref{figure:qualitative_study}), which makes $\Phi_{\text{c}}$ relatively local.

\subsection{Quantitative evaluation on ImageNet-1K}

\smallbreak\noindent\textbf{Ablation study on main components.} 
In Tab.~\ref{table:ablation_main}, we analyze the impact of each component in PerViT, which is denoted as (a), where C-stem refers to convolutional patch embedding stem\footnote{We increase feature dimensions of the models (c) (without $\Phi_{\text{c}}$) and (f, h, i) (without C-stem) accordingly to make FLOPs comparable to the others (a-i) to ensure the accuracy drops are not simply due to lower FLOPs.}.
We observe that the proposed attention $\Phi_{\text{p}}$ brings consistent gains to models (b, f, g, i) with relative improvements of 1.4$\sim$4.2\%p for (a, d, e, h) respectively.
Among three main components ($\Phi_{\text{p}}$, C-stem, CPE), $\Phi_{\text{p}}$ has the most significant impact on PerViT (a), losing 1.5\%p Top-1 accuracy without $\Phi_{\text{p}}$, \ie, model (b).
Surprisingly, PerViT without content-based attention $\Phi_{\text{c}}$, model (c), achieves decent performance, almost equalling to the accuracy of PerViT without position-based attention $\Phi_{\text{p}}$, model (b) (-0.5\%p).
The results verify that the proposed peripheral attention, which achieves comparable level of efficacy to the content-based attention, learns to generate reliable spatial attentions for visual recognition.
We also implement the proposed position-based attention $\Phi_{\text{p}}$ on DeiT~\cite{touvron2021deit} baseline and compare the results with recent state-of-the-art RPE methods.
As seen in Tab.~\ref{table:rpe_comparison}, the large improvements over the previous RPE methods~\cite{chu2021cpvt, wu2021irpe} further verify the efficacy of the proposed peripheral position encoding (PPE).
To confirm that the impact of $\Phi_{\text{p}}$ is consistent with large models, we conduct similar ablations using PerViT-S/M in Tab.~\ref{table:ablation_pervit_tsm};
without $\Phi_{\text{p}}$, the accuracy consistently drops for all the three models.
Comparing (b) with (c), we observe that C-Stem and CPE are less effective for large models, bringing 1.3\%p and 0.1\%p gains for Small and Medium respectively whereas they improve the Tiny model by 5.1\%p.
In contrast, the impact of $\Phi_{\text{p}}$ is consistent across different model sizes, bringing ~1\%p gains for all the three models.
The better efficacy of $\Phi_{\text{p}}$ for larger models, we hypothesize, is due to its flexibility in modeling local/global spatial attentions while C-Stem/CPE are designed only to be local.

\smallbreak\noindent\textbf{Sample-efficiency of PerViT.}
To investigate the training sample efficiency of our model, we train PerViT-S with ImageNet subsampled by fractions of 50\% and 25\% and evaluate it on full-sized test set of ImageNet-1K. For each subsamples, we increase the number of epochs to present models with a fixed number of images. Table~\ref{table:subsampling_experiments} compare our results with DeiT~\cite{touvron2021deit};
our model consistently surpasses the baseline for all subsampled datasets, showing its robustness under limited training data.

\begin{table}[t]
	\begin{minipage}{0.6\linewidth}
		\centering
        \caption{Study on the effect of each component in PerViT.}
        \vspace{1.0mm}
        \scalebox{0.73}{
            \begin{tabular}{cccccccc}
            
            \toprule
            
            Reference & $\Phi_{\text{p}}$ & $\Phi_{\text{c}}$ & C-stem & CPE & Top-1 & Top-5 & FLOPs (G) \\
            
            \midrule
            
            (a) & \cmark  & \cmark  & \cmark  & \cmark & 78.8 & 94.3 & 1.6 \\
            (b) & \xmark  & \cmark  & \cmark  & \cmark & 77.3 & 94.1 & 1.6 \\
            (c) & \cmark  & \xmark  & \cmark  & \cmark & 76.8 & 93.5 & 1.6 \\
            (d) & \cmark  & \cmark  & \xmark  & \cmark & 77.8 & 94.0 & 1.5 \\
            (e) & \cmark  & \cmark  & \cmark  & \xmark & 78.1 & 94.0 & 1.6 \\
            (f) & \xmark  & \cmark  & \xmark  & \cmark & 76.3 & 93.2 & 1.5 \\
            (g) & \xmark  & \cmark  & \cmark  & \xmark & 76.7 & 93.3 & 1.6 \\
            (h) & \cmark  & \cmark  & \xmark  & \xmark & 76.5 & 93.4 & 1.5 \\
            (i) & \xmark  & \cmark  & \xmark  & \xmark & 72.3 & 93.4 & 1.5 \\
            
            \bottomrule
            
            \end{tabular}
        }
        \vspace{1.0mm}
        \label{table:ablation_main}
	\end{minipage}\hfill
	\begin{minipage}{0.35\linewidth}
		\centering
        \caption{Comparisons between different relative position encodings with DeiT-Tiny~\cite{touvron2021deit} as a baseline.}
        \vspace{2.0mm}
        \scalebox{0.83}{
            \begin{tabular}{ccc}
            
            \toprule
            
            \multirow{2}{*}{Method}  & \multirow{2}{*}{Top-1} & \multirow{2}{*}{\shortstack{FLOPS\\(G)}} \\
            & & \\
            \midrule
            DeiT-T~\cite{touvron2021deit}    & 72.2 & 1.3 \\
            + CPVT~\cite{chu2021cpvt} & 73.4 & 2.1 \\
            + iRPE~\cite{wu2021irpe}  & 73.7 & 1.1 \\
            \textbf{+ PPE (ours)}     & 74.4 & 1.1 \\
            \bottomrule
            
            \end{tabular}
        }
        \vspace{3.0mm}
        \label{table:rpe_comparison}
	\end{minipage}
	\vspace{-4.0mm}
\end{table}

\begin{table}

	\begin{minipage}{0.4\linewidth}
		\centering
        \caption{Ablation on PerViT-T/S/M: the effects of $\Phi_{\text{p}}$, C-Stem, and CPE.}
		\vspace{2.0mm}
        \scalebox{0.75}{
            \begin{tabular}{lccccc}
            
            \toprule
            \multirow{2}{*}{Ref.} & \multirow{2}{*}{$\Phi_{\text{p}}$} & \multirow{2}{*}{\shortstack{C-Stem\\\& CPE}} & \multirow{2}{*}{T} & \multirow{2}{*}{S} & \multirow{2}{*}{M} \\
            & & & & & \\
            \midrule
            
            (a) & \cmark & \cmark & 78.8 & 82.1 & 82.9 \\
            (b) & \xmark & \cmark & 77.3 & 81.1 & 81.9 \\
            (c) & \xmark & \xmark & 72.2 & 79.8 & 81.8 \\
                
            \bottomrule
            
            \end{tabular}
        }
        \vspace{0.0mm}
        \label{table:ablation_pervit_tsm}
	\end{minipage}\hfill
	\begin{minipage}{0.55\linewidth}
		\centering
		\caption{Top-1 accuracy comparisons with DeiT-S~\cite{touvron2021deit} under different subsampling ratios: \{100\%, 50\%, 25\%\}.}
		\vspace{2.0mm}
        \scalebox{0.8}{
		\begin{tabular}{ccccc}
			\toprule
			
			\multirow{2}{*}{\shortstack{Subsampling\\ratio}} & \multicolumn{2}{c}{Top-1} & \multicolumn{2}{c}{Top-5} \\
			& DeiT-S & PerViT-S & DeiT-S & PerViT-S \\
			
			\midrule
			
			100\% & 79.9 & 82.1 & 95.0 & 95.8 \\
			50\% & 74.6 & 77.4 & 91.8 & 93.1 \\
			25\% & 61.8 & 67.5 & 82.6 & 86.9 \\
			
			\bottomrule
		\end{tabular}
		}
		\vspace{2.0mm}
		\label{table:subsampling_experiments}
	\end{minipage}	
	\vspace{-1.0mm}
\end{table}

\begin{table}[t]
	\begin{minipage}{0.3\linewidth}
		\centering
		\caption{Ablation study on different initialization methods (top section) and network designs (bottom section) for the position-based attention $\Phi_{\text{p}}$.}
		\vspace{4.0mm}
        \scalebox{0.72}{
		\begin{tabular}{lcc}
			\toprule
			\multirow{2}{*}{\shortstack{Init. method\\for $\Phi_{p}$}}     & \multirow{2}{*}{Top-1} & \multirow{2}{*}{Top-5} \\
			& & \\
			\midrule
			\textbf{Peripheral}       & \multirow{2}{*}{78.8} & \multirow{2}{*}{94.3} \\
			\textbf{(ours)} & & \\
			Conv             & 78.6 & 93.8 \\
			Rand             & 78.5 & 93.6 \\
			\midrule\midrule
			\multirow{2}{*}{\shortstack{Network\\design for $\Phi_{\text{p}}$}}    & \multirow{2}{*}{Top-1} & \multirow{2}{*}{Top-5} \\
			& & \\
			\midrule
			\textbf{Euc} + $\mathcal{N}$ + \textbf{ML}  & \multirow{2}{*}{78.8} & \multirow{2}{*}{94.3} \\
			\textbf{(ours)} & & \\
			Euc + ML                  & 77.9 & 94.0 \\
			Euc                       & 78.0 & 94.0 \\
			Lrn + $\mathcal{N}$ + ML  & 77.8 & 94.0 \\
			Lrn + ML                  & 77.5 & 93.8 \\
			Lrn                       & 77.6 & 93.8 \\
			
			\bottomrule
		\end{tabular}
		}
		\vspace{2.0mm}
		\label{table:phip_study}
	\end{minipage}
	\hfill
	\begin{minipage}{0.65\linewidth}
		\centering
        \caption{Model performance on ImageNet-1K~\cite{deng2009imagenet}.}
		\vspace{2.0mm}
        \scalebox{0.75}{
            \begin{tabular}{clccc}
            
            \toprule
            \multicolumn{2}{c}{Model} & Size (M) & FLOPs (G) & Top-1 (\%) \\
                
            \midrule\multirow{8}{*}{\shortstack{Pyramidal\\Vision\\Transformers\\ \ \\ \ \\({\em multi-resolution})}} 
            & PVT-T~\cite{wang2021pvt}                       & 13  & 1.9 & 75.1 \\
            & CoaT-Lite-T~\cite{xu2021coat}                  & 5.7 & 1.6 & 77.5 \\
            \cline{2-5} \\[-2.0ex]
            
            & Swin-T~\cite{liu2021swin}                      & 28  & 4.5 & 81.3 \\
            & CoaT-Lite-S~\cite{xu2021coat}                  & 20  & 4.0 & 81.9  \\
            & Focal-T~\cite{yang2021focal}                   & 29  & 4.9 & 82.2 \\
            \cline{2-5} \\[-2.0ex]
            
            & Swin-S~\cite{liu2021swin}                      & 50  & 8.7 & 83.0 \\
            & CoaT-Lite-M~\cite{xu2021coat}                  & 45  & 9.8 & 83.6 \\
            & Focal-S~\cite{yang2021focal}                   & 51  & 9.1 & 83.5 \\
            
            \midrule\multirow{11}{*}{\shortstack{Columnar\\Vision\\Transformers\\ \ \\ \ \\({\em single-resolution})}} 
            & DeiT-T~\cite{touvron2021deit}                 & 5.7 & 1.3 & 72.2 \\
            & XCiT-T12/16~\cite{el2021xcit}                 & 7.0 & 1.2 & 77.1 \\
            & \textbf{PerViT-T (ours)}                      & 7.6 & 1.6 & 78.8 \\
            \cline{2-5} \\[-2.0ex]
            
            & DeiT-S~\cite{touvron2021deit}                 & 22  & 4.6 & 79.8 \\
            & T2T-ViT$_{t}$-14~\cite{yuan2021t2tvit}        & 22  & 6.1 & 81.7 \\
            & XCiT-S12/16~\cite{el2021xcit}                 & 26  & 4.8 & 82.0 \\
            & \textbf{PerViT-S (ours)}                      & 21  & 4.4 & 82.1 \\
            \cline{2-5} \\[-2.0ex]
            
            & DeiT-B~\cite{touvron2021deit}                 & 86  & 18  & 81.8 \\
            & T2T-ViT$_{t}$-24~\cite{yuan2021t2tvit}        & 64  & 15  & 82.6 \\
            & XCiT-S24/16~\cite{el2021xcit}                 & 48  & 9.1 & 82.6 \\
            & \textbf{PerViT-M (ours)}                      & 44  & 9.0 & 82.9 \\
                
            \bottomrule
            
            \end{tabular}
        }
        \vspace{0.0mm}
        \label{table:sota_comparison}
	\end{minipage}
\end{table}

\smallbreak\noindent\textbf{Ablation study on $\Phi_{\text{p}}$.} 
The top section of Tab.~\ref{table:phip_study} reports results of PerViT-T with different parameter initialization methods for $\Phi_{\text{p}}$ where peripheral denotes the proposed peripheral initialization, conv refers to convolutional initialization such that $s_{l} = -5.0$ and $v_{l} = 3.0$ for all $l \in [N_l]$, and rand refers to random initialization for all parameters in $\Phi_{\text{p}}$: $w_r$, $\mathbf{W}^{(l)}_{\text{p1}}$, $\mathbf{W}^{(l,h)}_{\text{p2}}$, $\pmb{\gamma}_{\text{p1}}^{(l)}$, $\pmb{\beta}_{\text{p1}}^{(l)}$, ${\gamma}_{\text{p2}}^{(l, h)}$, and ${\beta}_{\text{p2}}^{(l, h)}$.
The results show the efficacy of our peripheral initialization which is also supported by the results in Fig.~\ref{fig:tiny_weight} and~\ref{fig:nonlocality}: $\Phi_{\text{p}}$ provides early local and late global attentions, suggesting that peripheral initialization effectively reduces burden in learning such form of attentions.
The bottom section of Tab.~\ref{table:phip_study} studies network designs for $\Phi_{\text{p}}$ where $\mathcal{N}$ represents the proposed peripheral projection, \ie, projecting input distance representation by referring neighbors $\mathcal{N}$, ML refers to multi-layer design of $\Phi_{\text{p}}$, and Euc \& Lrn indicate the type of embedding $\mathbf{R}$: Euc is relative Euclidean distances ($\mathbf{R}_{\mathbf{q}, \mathbf{k}, :} = \text{concat}_{r \in [D_r]}[w_r \cdot \mathbf{R}]$) where Lrn is relative distances between learnable vectors ($\mathbf{R} \in \mathbb{R}^{HW \times HW \times D_r}$).
Without $\mathcal{N}$ and ML, we observe consistent accuracy drops for Euc and Lrn by 0.8\%p and 0.2\%p respectively.
A sole multi-layer projection hardly improves accuracy but the model performs the best when $\mathcal{N}$ is jointly used, meaning that both need to complement each other to provide diverse attention shapes as in Fig~\ref{fig:tiny_weight}.
Furthermore, Euc models consistently surpasses Lrn models, implying the Euclidean distance is more straightforward encoding type than learnable vectors in capturing spatial configurations of images.

\smallbreak\noindent\textbf{Comparison with state of the arts.} 
Table~\ref{table:sota_comparison} summarizes the results of our method and recent state of the arts.
For fair comparison, the baselines used in our comparison are trained using $224 \times 224$ input resolution without distillations, and are grouped into either pyramidal or columnar ViT based on the network designs, \ie, multi- or single-resolution feature processing, where the results are partitioned according to model sizes within each group.
As shown in the bottom section of Tab~\ref{table:sota_comparison}, the proposed method achieves consistent improvements over the recent columnar ViT methods~\cite{d2021convit, el2021xcit, touvron2021deit, yuan2021t2tvit} while showing competitive results to the pyramidal counterparts.
Refer to the Appendix~\ref{sec:supp_results_analyses} for additional experimental results and analyses.


\section{Scope and Limitations}
\label{sec:limitation}
Despite the interpretability and effectiveness of PerViT, it still leaves much room for improvements.
First, PerViT-Attention (Eq.~\ref{eq:peripheral-attention}) is based on the original self-attention formulation~\cite{vaswani2018attention}, thus directly inheriting its limitations~\cite{dosovitskiy2021vit, touvron2021deit}, \eg, quadratic complexity w.r.t. input resolution.
The computational efficiency could be further improved by approximating low-rank matrices as in~\cite{choromanski2021performer, li2022uniformer, xu2021coat}.
Second, given the ability to process high-resolution input with feasible complexity, the efficacy of PerViT could be improved by adopting multi-resolution pyramidal design following recent trend of ViT designs~\cite{dong2021cswin, lee2022mpvit, li2022uniformer, liu2021swin, wang2021pvt, wu2021cvt, xu2021coat, yang2021focal}.
Third, the focus of this paper is model development \& exploration for image classification task but we believe the proposed idea is broadly generalizable to other vision applications such as object detection and segmentation.
We leave this to future work.

\section{Conclusion}
This paper explores blending human peripheral vision with machine vision for effective visual recognition, and introduces Peripheral Vision Transformer which learns to provide diverse position-based attentions to model peripheral vision using peripheral projections and initialization.
We have systematically investigated the inner workings of the proposed network and observed that the network enjoys the benefits of both convolution and self-attention by learning to decide level of the locality and dynamicity for the feature transformations, by the network itself given training data.
The consistent improvements over the baseline models on ImageNet-1K classification across different model sizes and in-depth ablation study confirm the efficacy of the proposed approach.

\clearpage

\section{Acknowledgments and Disclosure of Funding}
This work was supported by the IITP grants (No.2021-0-01696: High-Potential Individuals Global Training
Program (40\%),  No.2022-0-00290: Visual Intelligence for Space-Time Understanding and Generation based on Multi-layered Visual Common Sense (50\%), No.2019-0-01906: AI Graduate School Program -
POSTECH (10\%)) funded by Ministry of Science and ICT, Korea. This work was done while Juhong Min was working as an intern at Microsoft Research Asia.

\appendix

\newcommand{\beginsupplement}{%
        \setcounter{table}{0}
        \renewcommand{\thetable}{A\arabic{table}}%
        \setcounter{figure}{0}
        \renewcommand{\thefigure}{A\arabic{figure}}%
     }
\beginsupplement


\smallbreak\smallbreak\smallbreak\smallbreak
\textbf{{\LARGE Appendix}}
\smallbreak\smallbreak

In the Appendix, we provide additional results and analyses of the proposed method, and implementation details.
In Sec.~\ref{sec:supp_peripheral_init}, we provide a complete derivation on the peripheral initialization introduced in Sec.~\ref{sec:mpa_main} of our main paper.
In Sec.~\ref{sec:supp_peripheral_region}, we present additional details on peripheral region classification presented in Sec~\ref{sec:innerworking_main}.
In Sec.~\ref{sec:supp_mpa_conv_mhsa}, we prove that an Multi-head Peripheral Attention (MPA) in two extreme cases of semi-dynamic attention and position bias injection is in turn convolution and multi-head self-attnetion layers respectively.
In Sec.~\ref{sec:supp_results_analyses}, we provide in-depth analyses of inner workings of PerViT and recent baselines of~\cite{touvron2021deit, wu2021irpe} both quantitatively and qualitatively.
In Sec.~\ref{sec:supp_layout_details} and \ref{sec:supp_training_details}, we provide network layouts of different model sizes, training hyperparameters, and implementation details.
We conclude this paper with a short discussion on potential impacts of our work in Sec.~\ref{sec:supp_broader_impacts}.
We made the code and data publicly available\footnote{Code: \url{https://github.com/juhongm999/pervit}}\blfootnote{Project page: \url{http://cvlab.postech.ac.kr/research/PerViT/}}.

\smallbreak

\section{Peripheral Initialization with a Complete Derivation}
\label{sec:supp_peripheral_init}
Recall the definition of the peripheral position encoding introduced in Sec.~\ref{sec:mpa_main}:
\begin{align}
    \mathbf{R}' &\coloneqq \text{ReLU} \left( \text{IN} ( \text{PP} ( \mathbf{R}; \mathbf{W}_{\text{p1}} ); \pmb{\gamma}_{\text{p1}}, \pmb{\beta}_{\text{p1}} )\right), \\ \Phi^{(h)}_{\text{p}} (\mathbf{R}) &\coloneqq \sigma \left( \text{IN} ( \text{PP} ( \mathbf{R}'; \mathbf{W}_{\text{p2}}^{(h)} ); {\gamma}^{(h)}_{\text{p2}}, {\beta}^{(h)}_{\text{p2}} ) \right),
\end{align}
where $\mathbf{R}_{\mathbf{q}, \mathbf{k}, :} \coloneqq \text{concat}_{r \in [D_r]}[w_r \cdot \mathbf{R}_{\mathbf{q}, \mathbf{k}}^{\text{euc}}]$ is Euclidean distances between query $\mathbf{q}$ and key $\mathbf{k}$ position in $D_{r}$ different scales, $\pmb{\gamma}_{\text{p1}}^{(h)}, \pmb{\beta}_{\text{p1}}^{(h)} \in \mathbb{R}^{D_{\text{hid}}}$, and $\gamma_{\text{p2}}^{(h)}, \beta_{\text{p2}}^{(h)} \in \mathbb{R}$ are weights/biases of the instance normalization layers~\cite{ulyanov2017instancenorm} $\text{IN}(\cdot)$, $\mathbf{W}_{\text{p1}} \in \mathbb{R}^{K^2 \times D_{\text{r}} \times D_{\text{hid}}}$ and $\mathbf{W}_{\text{p2}}^{(h)} \in \mathbb{R}^{K^2 \times D_{\text{hid}}}$ are learnable parameters of the peripheral projections $\text{PP}(\cdot)$, and $\sigma(\cdot)$ is a sigmoid function.
Specifically, the peripheral projection with parameter $\mathbf{W} \in \mathbb{R}^{K^2 \times D_{\text{in}} \times D_{\text{out}}}$ transforms the input $\mathbf{R} \in \mathbb{R}^{HW \times HW \times D_{\text{in}}}$ by referring neighboring distance representations in the key dimension as follows:
\begin{align}
    \text{PP}(\mathbf{R}; \mathbf{W})_{\mathbf{q}, \mathbf{k}, :} \coloneqq \sum_{\mathbf{n} \in \mathcal{N}(\mathbf{k})} \mathbf{R}_{\mathbf{q}, \mathbf{n}, :} \mathbf{W}_{\mathbf{n}-\mathbf{k}, :, :},
\end{align}
where $\mathcal{N}(\mathbf{k})$ returns $K^2 $ neighboring positions around position $\mathbf{k}$ including itself.

Now assume we have
\begin{align}
    \label{eq:periinit_param}
    w_r \coloneqq -c_{1}, \ \ \ \ \ \ \mathbf{W}_{\text{p1}} \coloneqq c_{2}J_{K^{2}, D_{\text{r}}, D_{\text{hid}}}, \ \ \ \ \ \ \mathbf{W}^{(h)}_{\text{p2}} \coloneqq c_{2}J_{K^{2}, D_{\text{hid}}}, \ \ \ \ \ \ \pmb{\gamma}_{\text{p1}} \coloneqq \mathbf{1}_{D_{\text{hid}}}, \ \ \ \ \ \ \pmb{\beta}_{\text{p1}} \coloneqq \mathbf{0}_{D_{\text{hid}}},
\end{align}
for all $r \in [D_{\text{r}}]$, and $h \in [N_h]$ where $c_1, c_2 \in \mathbb{R}^{+}$ are positive reals, $J_{N, M} \in \mathbb{R}^{N \times M}$ refers to all-one matrix in size $N \times M$, $\mathbf{1}_{D_{\text{hid}}} = [1, ..., 1]^{\top} \in \mathbb{R}^{D_{\text{hid}}}$, and $\mathbf{0}_{D_{\text{hid}}} \in \mathbb{R}^{D_{\text{hid}}}$ is a zero vector.
The parameterizations in Eq.~\ref{eq:periinit_param} are applied for all the layers in the network ($l \in [N_l]$) but we omit the layer indices in this derivation for brevity.
Next, we show that (1) this initialization of these learnable parameters in $\Phi_{\text{p}}^{(h)}$ provides local attentions for every layer in the beginning of the training stage and (2) the respective size and strength of the local attentions are controlled by the bias $\beta_{\text{p2}}^{(h)}$ and weight $\gamma_{\text{p2}}^{(h)}$ of the second instance norm.

\clearpage

\smallbreak\noindent\textbf{Step 1.} 
Our first step is to prove the parameterization (Eq.~\ref{eq:periinit_param}) provides local attention after the second peripheral projection, \ie, $\text{PP}(\mathbf{R}'; \mathbf{W}_{\text{p2}}^{(h)} )_{\mathbf{q}, \mathbf{k}_{i}, :} > \text{PP}(\mathbf{R}'; \mathbf{W}_{\text{p2}}^{(h)})_{\mathbf{q}, \mathbf{k}_{j}, :}$ given $\mathbf{R}^{\text{euc}}_{\mathbf{q}, \mathbf{k}_i} < \mathbf{R}^{\text{euc}}_{\mathbf{q}, \mathbf{k}_j}$.
Consider the first peripheral projection given $\mathbf{R} \in \mathbb{R}^{HW \times HW \times D_{\text{r }}}$:
\begin{align}
    \text{PP}(\mathbf{R}; \mathbf{W}_{\text{p1}})_{\mathbf{q}, \mathbf{k}, :} &= \sum_{\mathbf{n} \in \mathcal{N}(\mathbf{k})} \mathbf{R}_{\mathbf{q}, \mathbf{n}, :} \mathbf{W}_{\text{p1} \ \mathbf{n} - \mathbf{k}, :, :} \\
    &= c_{2}\sum_{\mathbf{n} \in \mathcal{N}(\mathbf{k})} \mathbf{R}_{\mathbf{q}, \mathbf{n}, :} \cdot J_{D_{\text{r}}, D_{\text{hid}}} \\
    &= c_{2} \sum_{\mathbf{n} \in \mathcal{N}(\mathbf{k})} \left( \sum_{r \in [D_{\text{r}}]} \mathbf{R}_{\mathbf{q}, \mathbf{n}, r} \right) \mathbf{1}_{D_{\text{hid}}} \\
    &= c_{2} \sum_{\mathbf{n} \in \mathcal{N}(\mathbf{k})} \left( \sum_{r \in [D_{\text{r}}]} w_{\text{r}} \cdot \mathbf{R}_{\mathbf{q}, \mathbf{n}}^{\text{euc}} \right) \mathbf{1}_{D_{\text{hid}}} \\
    \label{eq:pp1}
    &= -c_{1} c_{2} D_{\text{r}} \left( \sum_{\mathbf{n} \in \mathcal{N}(\mathbf{k})} \mathbf{R}_{\mathbf{q}, \mathbf{n}}^{\text{euc}} \right) \mathbf{1}_{D_{\text{hid}}},
\end{align}
which increases channel dimension to $D_{\text{hid}}$ and scales the Euclidean distance $\mathbf{R}_{\mathbf{q}, \mathbf{k}}^{\text{euc}}$ by $-c_{1}c_{2}D_{r}$ after summing its neighbors.
Note that the negation in Eq.~\ref{eq:pp1} gives the inequality of
\begin{align}
    \text{PP}(\mathbf{R}; \mathbf{W}_{\text{p1}})_{\mathbf{q}, \mathbf{k}_i, :} > \text{PP}(\mathbf{R}; \mathbf{W}_{\text{p1}})_{\mathbf{q}, \mathbf{k}_j, :},
\end{align}
given $\mathbf{R}^{\text{euc}}_{\mathbf{q}, \mathbf{k}_i} < \mathbf{R}^{\text{euc}}_{\mathbf{q}, \mathbf{k}_j}$\footnote{Note that $\sum_{\mathbf{n} \in \mathcal{N}(\mathbf{k}_i)} \mathbf{R}_{\mathbf{q}, \mathbf{n}}^{\text{euc}} < \sum_{\mathbf{n} \in \mathcal{N}(\mathbf{k}_j)} \mathbf{R}_{\mathbf{q}, \mathbf{n}}^{\text{euc}}$ as we deal with Euclidean distances.}.
Given $\mathbf{R}^{\text{PP1}} \coloneqq \text{PP}(\mathbf{R}; \mathbf{W}_{\text{p1}}) \in \mathbb{R}^{HW \times HW \times D_{\text{hid}}}$, the first instance norm $\text{IN}(\cdot; \mathbf{1}_{D_{\text{hid}}}, \mathbf{0}_{D_{\text{hid}}})$ provides normalized output as follows:
\begin{align}
    \text{IN}(\mathbf{R}^{\text{PP1}}; \mathbf{1}_{D_{\text{hid}}}, \mathbf{0}_{D_{\text{hid}}})_{\mathbf{q}, \mathbf{k}, c} \coloneqq \frac{\mathbf{R}^{\text{PP1}}_{\mathbf{q}, \mathbf{k}, c} -  \EX_{\mathbf{m} \sim \mathcal{P}} \left[ \mathbf{R}^{\text{PP1}}_{\mathbf{q}, \mathbf{m}, c} \right] }{  \EX_{\mathbf{n} \sim \mathcal{P}} \left[ \left( \mathbf{R}^{\text{PP1}}_{\mathbf{q}, \mathbf{n}, c} -  \EX_{\mathbf{m} \sim \mathcal{P}} \left[ \mathbf{R}^{\text{PP1}}_{\mathbf{q}, \mathbf{m}, c} \right] \right)^{2} \right]   } \cdot \mathbf{1}_{D_{\text{hid}} \ c} + \mathbf{0}_{D_{\text{hid}} \ c},
\end{align}
which simply normalizes each channel dimension to unit Gaussian while preserving the inequality, \ie, $\text{IN}(\mathbf{R}^{\text{PP1}})_{\mathbf{q}, \mathbf{k}_{i}, :} > \text{IN}(\mathbf{R}^{\text{PP1}})_{\mathbf{q}, \mathbf{k}_{j}, :}$ for all $\mathbf{q}, \mathbf{k}_{i}, \mathbf{k}_{j} \in \mathcal{P}$.
The ReLU nonlinearlity takes the normalized output and suppresses negative activations in $\mathbf{R}^{\text{IN1}} \coloneqq \text{IN}(\mathbf{R}^{\text{PP1}}) \in \mathbb{R}^{HW \times HW \times D_{\text{hid}}}$, keeping the inequality only for positive activations:
\begin{align}
    \mathbf{R}'_{\mathbf{q}, \mathbf{k}_i, :} = \text{ReLU}\left( \mathbf{R}^{\text{IN1}} \right)_{\mathbf{q}, \mathbf{k}_i, :} > \text{ReLU}\left( \mathbf{R}^{\text{IN1}} \right)_{\mathbf{q}, \mathbf{k}_j, :} = \mathbf{R}'_{\mathbf{q}, \mathbf{k}_j, :},
\end{align}
for $\mathbf{q}, \mathbf{k}_i, \mathbf{k}_j \in \mathcal{P}$ that satisfy $\mathbf{R}^{\text{IN1}}_{\mathbf{q}, \mathbf{k}_i, :}, \mathbf{R}^{\text{IN1}}_{\mathbf{q}, \mathbf{k}_j, :} > \mathbf{0}_{D_{\text{hid}}}$.
Now consider second peripheral projection:
\begin{align}
    \mathbf{R}^{\text{PP2}}_{\mathbf{q}, \mathbf{k}} &= \text{PP}(\mathbf{R}'; \mathbf{W}^{(h)}_{\text{p2}})_{\mathbf{q}, \mathbf{k}} \\
    &= \sum_{\mathbf{n} \in \mathcal{N}(\mathbf{k})} \mathbf{R}'_{\mathbf{q}, \mathbf{n}, :} \mathbf{W}^{(h)}_{\text{p2} \ \mathbf{n} - \mathbf{k}, :} \\
    &= c_{2}\sum_{\mathbf{n} \in \mathcal{N}(\mathbf{k})} \mathbf{R}'_{\mathbf{q}, \mathbf{n}, :} \cdot \mathbf{1}_{D_{\text{hid}}} \\
    &= c_{2} \sum_{\mathbf{n} \in \mathcal{N}(\mathbf{k})} \left( \sum_{h \in [D_{\text{hid}}]} \mathbf{R}'_{\mathbf{q}, \mathbf{n}, h} \right) \\
    \label{eq:pp2}
    &= c_{2}D_{\text{hid}} \sum_{\mathbf{n} \in \mathcal{N}(\mathbf{k})} \mathbf{R}'_{\mathbf{q}, \mathbf{n}, d} 
\end{align}
for any $d \in [D_{\text{hid}}]$.
Similarly to the first $\text{PP}(\cdot)$, Eq.~\ref{eq:pp2} scales the input by $c_{2}D_{\text{hid}} \in \mathbb{R}^{+}$ after summing its neighbors.
Hence, $\mathbf{R}^{\text{PP2}}_{\mathbf{q}, \mathbf{k}_i} > \mathbf{R}^{\text{PP2}}_{\mathbf{q}, \mathbf{k}_j}$ holds given $\mathbf{R}^{\text{euc}}_{\mathbf{q}, \mathbf{k}_i} < \mathbf{R}^{\text{euc}}_{\mathbf{q}, \mathbf{k}_j}$ if $\mathbf{R}^{\text{PP2}}_{\mathbf{q}, \mathbf{k}_i}, \mathbf{R}^{\text{PP2}}_{\mathbf{q}, \mathbf{k}_j} > 0$, providing local attention. \QEDA

\begin{figure*}[t]
    \begin{center}

    \scalebox{0.38}{
    \centering
        \includegraphics{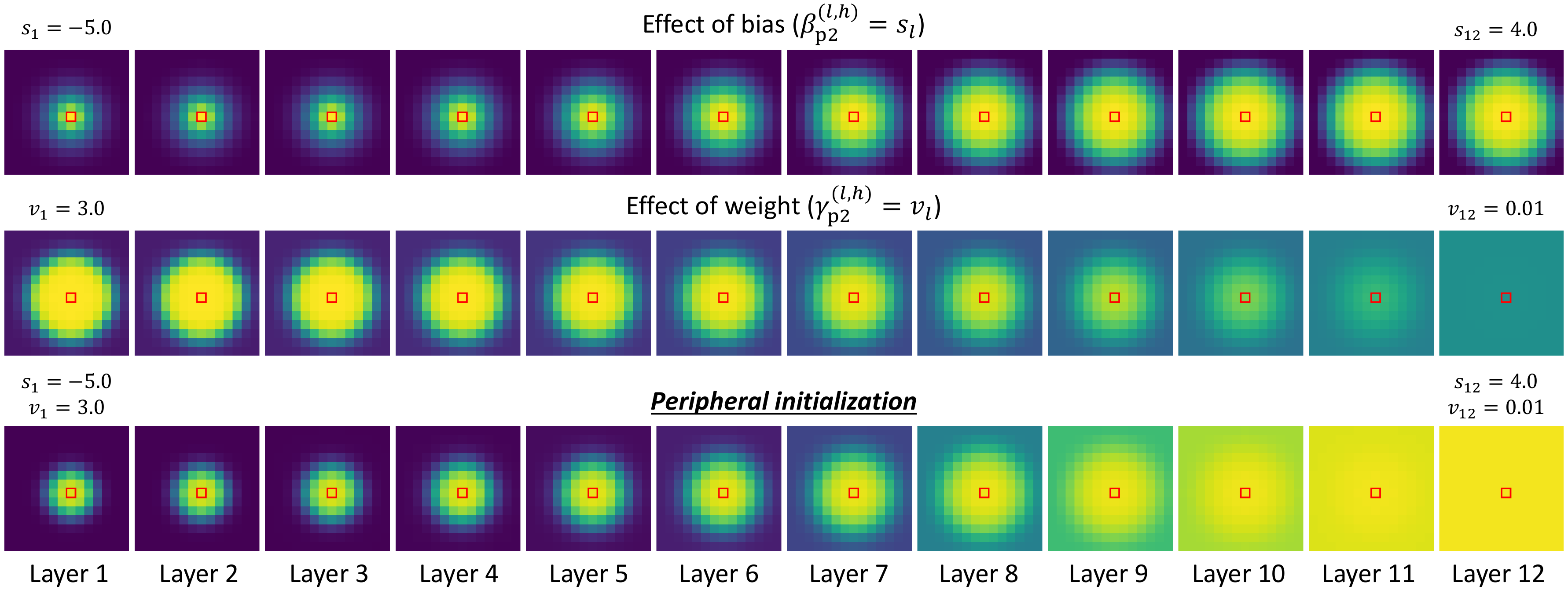}
    }
    \vspace{-0.5mm}
    \caption{Effects of the biases $\beta_{\text{p2}}^{(l,h)}$ (top) and weights $\gamma_{\text{p2}}^{(l,h)}$ (middle) to the size and strength of local attentions. Our network exploits early local and late global attentions in the beginning of the training stage, which are achieved via peripheral initialization (bottom).} 
    \vspace{-6.0mm}
    \label{fig:peri_init_sample}
    \end{center}   
\end{figure*}

\smallbreak\noindent\textbf{Step 2.} 
We now show that the respective size and strength of local attention in the peripheral position encoding $\Phi_{\text{p}}^{(h)}$ are controlled by the bias $\beta_{\text{p2}}^{(h)}$ and weight $\gamma_{\text{p2}}^{(h)}$ of the second instance norm, which is defined as follows:
\begin{align}
    \text{IN}(\mathbf{R}^{\text{PP2}}; \gamma_{\text{p2}}^{(h)}, \beta_{\text{p2}}^{(h)})_{\mathbf{q}, \mathbf{k}} &\coloneqq \frac{\mathbf{R}^{\text{PP2}}_{\mathbf{q}, \mathbf{k}} -  \EX_{\mathbf{m} \sim \mathcal{P}} \left[ \mathbf{R}^{\text{PP2}}_{\mathbf{q}, \mathbf{m}} \right] }{  \EX_{\mathbf{n} \sim \mathcal{P}} \left[ \left( \mathbf{R}^{\text{PP2}}_{\mathbf{q}, \mathbf{n}} -  \EX_{\mathbf{m} \sim \mathcal{P}} \left[ \mathbf{R}^{\text{PP2}}_{\mathbf{q}, \mathbf{m}} \right] \right)^{2} \right]   } \cdot \gamma_{\text{p2}}^{(h)} + \beta_{\text{p2}}^{(h)} \\
    \label{eq:periinit_final}
    &= \mathbf{R}^{\text{PP2-Norm}}_{\mathbf{q}, \mathbf{k}} \cdot \gamma_{\text{p2}}^{(h)} + \beta_{\text{p2}}^{(h)},
\end{align}
where $\mathbf{R}^{\text{PP2-Norm}}$ refers to unit-Gaussian normalized $\mathbf{R}^{\text{PP2}}$.
Note that the weight and bias terms in the above formulation (Eq.~\ref{eq:periinit_final}) have immediate control over the distribution of the position-based attention scores:
\begin{align}
    \Phi_{\text{p} \ \mathbf{q}, \mathbf{k}}^{(h)} = \sigma \left( \mathbf{R}^{\text{PP2-Norm}}_{\mathbf{q}, \mathbf{k}} \cdot \gamma_{\text{p2}}^{(h)} + \beta_{\text{p2}}^{(h)} \right).
\end{align}
For example, given some fixed query position $\mathbf{q}$, a large bias term encourages global attention:
\begin{align}
    \label{eq:beta_infty}
    \lim_{\beta_{\text{p2}}^{(h)}\to\infty} \Phi_{\text{p} \ \mathbf{q}, \mathbf{k}}^{(h)} = \lim_{\beta_{\text{p2}}^{(h)}\to\infty} \sigma \left( \mathbf{R}^{\text{PP2-Norm}}_{\mathbf{q}, \mathbf{k}} \cdot \gamma_{\text{p2}}^{(h)} + \beta_{\text{p2}}^{(h)} \right) = 1 \ \ \ \ \text{for all} \ \ \ \mathbf{k} \in \mathcal{P}, 
\end{align}
controlling the size of local attention as seen in the top row of Fig.~\ref{fig:peri_init_sample}, whereas small magnitude of the weight term makes attention distribution more uniform:
\begin{align}
    \lim_{\gamma_{\text{p2}}^{(h)}\to 0} \Phi_{\text{p} \ \mathbf{q}, \mathbf{k}}^{(h)} = \lim_{\gamma_{\text{p2}}^{(h)}\to 0} \sigma \left( \mathbf{R}^{\text{PP2-Norm}}_{\mathbf{q}, \mathbf{k}} \cdot \gamma_{\text{p2}}^{(h)} + \beta_{\text{p2}}^{(h)} \right) = \sigma \left( \beta_{\text{p2}}^{(h)} \right) \ \ \ \ \text{for all} \ \ \ \mathbf{k} \in \mathcal{P},
\end{align}
manipulating the strength of the local attention as seen in the middle row of Fig.~\ref{fig:peri_init_sample}. \QEDA

\smallbreak\noindent\textbf{Simulating peripheral initialization.}
Figure~\ref{fig:peri_init_sample} illustrates attention maps $\Phi_{\text{p} \ \mathbf{q}, :}^{(l, h)} \in \mathbb{R}^{HW}$ given a query position at the center, \ie, $\mathbf{q} = [7, 7]^{\top}$, under varying biases $\beta_{\text{p2}}^{(l,h)} = s_l$ and weights $\gamma_{\text{p2}}^{(l,h)} = v_l$ where the respective values are collected from uniform intervals: $s_{l} \in [-5.0, 4.0]$ and $v_l \in [3.0, 0.01]$ which satisfy $s_{l-1} < s_l$ and $v_{l-1} > v_l$.
The proposed peripheral initialization imposes strong locality in early layers ($s_1 = -5.0$ and $v_1 = 3.0$) and global attention ($s_{12} = 4.0$ and $v_{12} = 0.01$) in late layers, facilitating training of PerViT as demonstrated in Sec.~\ref{sec:innerworking_main} with Figs.~\ref{fig:tiny_weight} and \ref{fig:nonlocality}.

\clearpage

\section{Peripheral Region Classification}
\label{sec:supp_peripheral_region}
In Sec.~\ref{sec:innerworking_main} of the main paper, we have classified every feature transformation layer in PerViT into one of the four visual regions $\mathbb{P} \in \{\text{c}, \text{p}, \text{m}, \text{f}\}$ where the elements refer to central, para-central, mid, and far peripheral regions respectively.
We use the following formulation:
\begin{align}
    \label{eq:peri_classification}
    \text{PeripheralRegion}{(l,h)} \coloneqq \argmax_{p \in \mathbb{P}}\left[ \frac{1}{|\mathcal{P}|^{2}} \sum_{(\mathbf{q}, \mathbf{k}) \in \mathcal{P} \times \mathcal{P}} \Phi^{(l,h)}_{\text{p} \ \ \mathbf{q}, \mathbf{k}} \cdot \mathbbm{1}\left[\norm{\mathbf{q} - \mathbf{k}}_{2} \in \mathcal{I}_{p}\right] \right],
\end{align}
where we set $\mathcal{I}_{\text{c}} = [0, r_{\text{c}})$ , $\mathcal{I}_{\text{p}} = [r_{\text{c}}, r_{\text{p}})$, $\mathcal{I}_{\text{m}} = [r_{\text{p}}, r_{\text{m}})$, and $\mathcal{I}_{\text{f}} = [r_{\text{m}}, r_{\text{f}})$ in experiments.

\begin{figure*}[h]
    \begin{center}

    \scalebox{0.55}{
    \centering
        \includegraphics{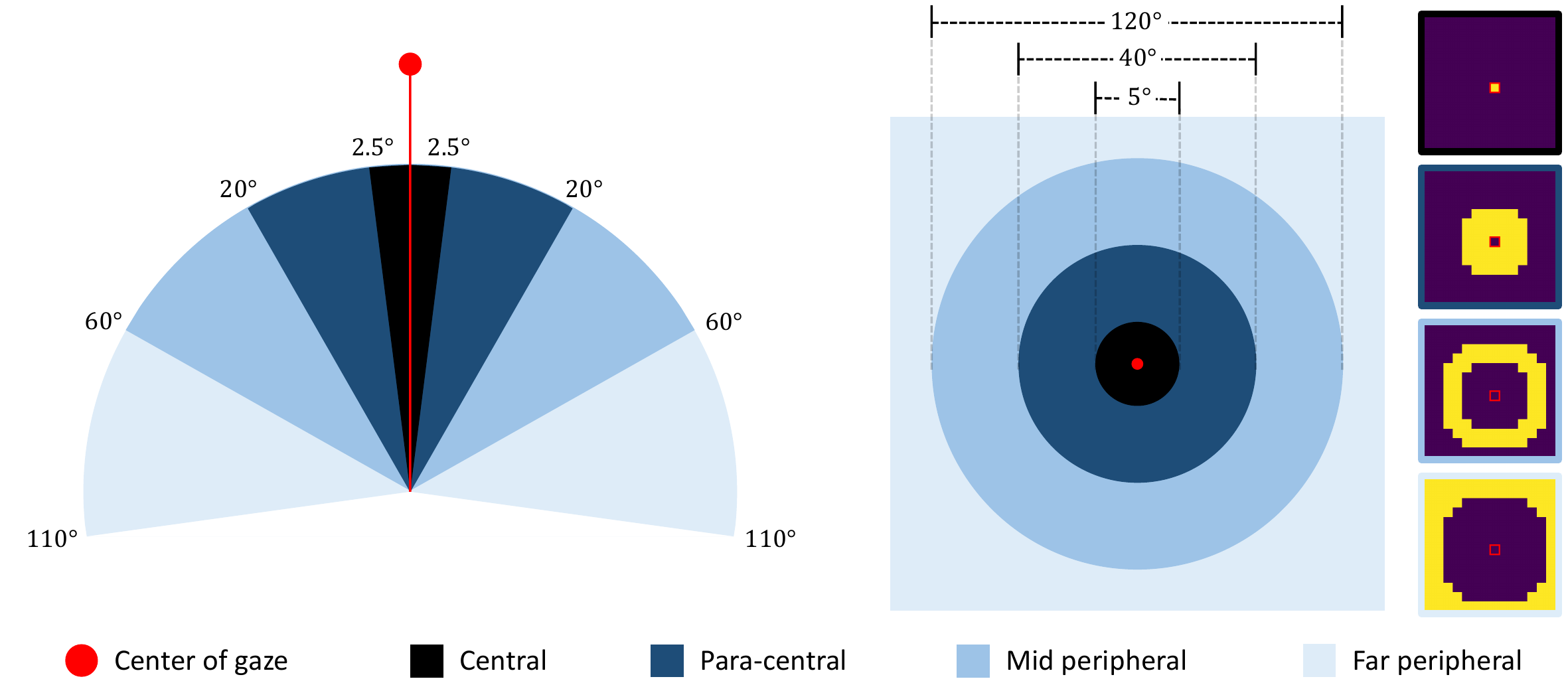}
    }
    \vspace{1.0mm}
    \caption{Peripheral vision of human eye (left). Peripheral vision of an MPA layer in PerViT (right).}
    \vspace{-3.0mm}
    \label{fig:angle}
    \end{center}   
\end{figure*}

According to vision science literature~\cite{hans2011peripheralvision}, \eg, physiology, ophthalmology, and optometry, the maximum extent of whole visual field is bounded within 100-110° (horizontal) angles from the gaze, providing 200-220° perceivable area like fan-shaped figure in the left of Fig.~\ref{fig:angle}.
The central vision, which we call central {\em region} in the context of attention map, refers to area corresponding 5° of whole visual field whereas the paracentral vision covers upto 8°.
The near-, mid-, and far-peripheral vision correspond to non-overlapping areas outside circles 8°, 60°, and 120° in diameter respectively.
When classifying the feature transformation layers in Sec.~\ref{sec:innerworking_main}, we adhere to these angular ratios to imitate human peripheral vision as faithfully as possible:
\begin{align}
    \label{eq:peri_ratio}
    \theta_{p} : 220^{\circ} = \pi r_{p}^{2} : HW,
\end{align}
where $\theta_{p} \in \{5^{\circ}, 40^{\circ}, 120^{\circ}, 220^{\circ}\}$ is a set of angles that divide visual field into four non-overlapping areas of central, para-central\footnote{We found that Eq.~\ref{eq:peri_ratio} gives radius of 1.5 given para-central angle of 8°, resulting in quite narrow interval: $[1.19, 1.5)$. Since the concept of paracentral (8°) and near-peripheral (60°) visions are used interchangeably in literature~\cite{enwiki:1044083194}, we set the angle for para-central region to some intermediary value between 8° and 60°, \ie, $\theta_{\text{p}} = 40^{\circ}$, so every peripheral region gets radii of (approximately) equal size as seen in the right of Fig.~\ref{fig:angle}.}, mid-, and far-peripheral regions respectively.
Solving the equality in Eq.~\ref{eq:peri_ratio} with respect to the radius $r_p$ gives
\begin{align}
    \label{eq:peri_radius}
    r_{p} = \sqrt{\frac{HW \cdot \theta_{p}}{220^{\circ}\pi}}.
\end{align}
We then have $r_{\text{c}} = 1.19$, $r_{\text{p}} = 3.37$, $r_{\text{m}} = 5.83$, and $r_{\text{f}} = 7.9$, \ie, $\mathcal{I}_{\text{c}} = [0, 1.19)$ , $\mathcal{I}_{\text{p}} = [1.19, 3.37)$, $\mathcal{I}_{\text{m}} = [3.37, 5.83)$, and $\mathcal{I}_{\text{f}} = [5.83, 7.9)$.
For the classification experiments performed in Fig.~\ref{fig:peri_cls} of the main paper, we classify two linear projections in an MLP and a $3 \times 3$ convolution in CPE as central regions as radii of their receptive fields approximately fall in the interval of $\mathcal{I}_{\text{c}} = [0, 1.19)$.

We consider each query location $\mathbf{q}$ of MPA, \eg, the position of a feature we want to transform, as a focal point, assuming each MPA in PerViT simultaneously processes $H \times W$ pixel locations with $H \times W$ different focal points given input feature size of $H \times W$. This assumption provides ring-shaped attentions if a query is located at the center of the feature map (Fig.~\ref{fig:angle}). While we have developed our narrative in the context of images (2D), the assumption deviates from the reality when considering FOV of a physical eyeball (3D).

\clearpage

\section{Proof of MPA as Convolution and MHSA}
\label{sec:supp_mpa_conv_mhsa}
In this section, we constructively prove that a Multi-head Peripheral Attention (MPA) layer in extreme case of semi-dynamic attention, \ie, strong attention at a small area, and position bias injection, \ie, relatively broad attention over the whole visual field, is in turn convolution and multi-head self-attention layers respectively.
We first recall the definition of the MPA:
\begin{align}
    \text{MPA}(\mathbf{X}) &\coloneqq \underset{h \in [N_h]}{\text{concat}} \big[\text{Peripheral-Attention}^{(h)}(\mathbf{X}, \mathbf{R}) \big] \mathbf{W}_{\text{out}} + \mathbf{b}_{\text{out}} \\
    &= \left( \sum_{h \in [N_h]} \text{Peripheral-Attention}^{(h)}(\mathbf{X}, \mathbf{R}) \right) \mathbf{W}^{(h)}_{\text{out}} + \mathbf{b}_{\text{out}},
\end{align}
where $\mathbf{W}^{(h)}_{\text{out}} = (\mathbf{W}_{\text{out}})_{(h-1) D_{h} + 1 : h  D_{h} + 1} \in \mathbb{R}^{D_{h} \times D_{\text{emb}}}$ and Peripheral-Attention is defined as
\begin{align}
    \text{Peripheral-Attention}^{(h)}(\mathbf{X}, \mathbf{R}) \coloneqq \text{Normalize}\left[\Phi^{(h)}_{\text{c}}(\mathbf{X}) \odot \Phi^{(h)}_{\text{p}}( \mathbf{R} ) \right] \mathbf{V}^{(h)}.
\end{align}

\smallbreak\noindent\textbf{MPA as a convolution.}
Assume the position-based function at each head is learned to perform `hard attention' on one of its surrounding positions, \ie, {\em an extreme semi-dynamic attention}.
To formally put, $\Phi_{\text{p}}^{(h)}(\mathbf{R} )_{\mathbf{q}, \mathbf{k}} \coloneqq \mathds{1}[0 = s(h) - \mathbf{R} _{\mathbf{q}, \mathbf{k}, :}] \in \mathbb{R}^{HW \times HW}$ where $\mathbf{R}  \in \mathbb{R}^{HW \times HW \times 2}$ is a matrix containing pair-wise offsets between query and key positions, \ie, $\mathbf{R} _{\mathbf{q}, \mathbf{k}} \coloneqq \mathbf{q} - \mathbf{k}$, and $s(h): [N_h] \xrightarrow{} \nabla_{k}$ is a bijective mapping of heads onto a fixed set of offsets
$\nabla_{k} = \{-\lfloor k/2 \rfloor, ..., \lfloor k/2 \rfloor\} \times \{-\lfloor k/2 \rfloor, ..., \lfloor k/2 \rfloor\}$, \ie, $N_h = k^2$.
Given the assumptions, consider Peripheral-Attention at query position $\mathbf{q}$:
\begin{align}
    \text{Peripheral-Attention}^{(h)} (\mathbf{X}, \mathbf{R})_{\mathbf{q}, :} &= \sum_{\mathbf{k} \in \mathcal{P}} \left( \frac{ \Phi^{(h)}_{\text{c}}(\mathbf{X})_{\mathbf{q}, \mathbf{k}} \cdot \Phi^{(h)}_{\text{p}}( \mathbf{R}  )_{\mathbf{q}, \mathbf{k}} }{ \sum_{\mathbf{j} \in \mathcal{P} }\Phi^{(h)}_{\text{c}}(\mathbf{X})_{\mathbf{q}, \mathbf{j}} \cdot \Phi^{(h)}_{\text{p}}( \mathbf{R}  )_{\mathbf{q}, \mathbf{j}} } \right) \mathbf{V}^{(h)}_{\mathbf{k}, :} \\
    &= \sum_{\mathbf{k} \in \mathcal{P}} \left( \frac{\Phi^{(h)}_{\text{c}}(\mathbf{X})_{\mathbf{q}, \mathbf{k}} \cdot \mathds{1}[s(h) = \mathbf{R} _{\mathbf{q}, \mathbf{k}, :} ] }{\sum_{\mathbf{j} \in \mathcal{P}} \Phi^{(h)}_{\text{c}}(\mathbf{X})_{\mathbf{q}, \mathbf{j}} \cdot \mathds{1}[s(h) = \mathbf{R} _{\mathbf{q}, \mathbf{j}, :} ] } \right) (\mathbf{X}\mathbf{W}^{(h)}_{\text{val}})_{\mathbf{k}, :} \\
    &= \sum_{\mathbf{k} \in \mathcal{P}} \left( \mathds{1}[s(h) = \mathbf{R} _{\mathbf{q}, \mathbf{k}, :} ] \right) (\mathbf{X}_{\mathbf{k}, :} \cdot \mathbf{W}^{(h)}_{\text{val}}) \\
    &= \left( \sum_{\mathbf{k} \in \mathcal{P}}  \mathds{1}[s(h) = \mathbf{q} - \mathbf{k} ]  \cdot \mathbf{X}_{\mathbf{k}, :} \right) \mathbf{W}^{(h)}_{\text{val}} \\
    &= \left( \sum_{\mathbf{k} \in \mathcal{P}}  \mathds{1}[\mathbf{k} = \mathbf{q} - s(h) ]  \cdot \mathbf{X}_{\mathbf{k}, :} \right) \mathbf{W}^{(h)}_{\text{val}} \\
    &= \mathbf{X}_{\mathbf{q} - s(h), :} \ \mathbf{W}^{(h)}_{\text{val}} \\
\end{align}
Assuming $D_h \geq D_{\text{emb}}$, the MPA is formulated as
\begin{align}
    \text{MPA}(\mathbf{X})_{\mathbf{q},:} &= \sum_{h \in [N_h]} \left( \text{Peripheral-Attention}^{(h)} (\mathbf{X}, \mathbf{R})_{\mathbf{q}, :} \right) \mathbf{W}^{(h)}_{\text{out}} + \mathbf{b}_{\text{out}} \\
    &= \sum_{h \in [N_h]} \left( \mathbf{X}_{\mathbf{q}-s(h),:} \mathbf{W}^{(h)}_{\text{val}} \right) \mathbf{W}^{(h)}_{\text{out}} + \mathbf{b}_{\text{out}} \\
    &= \sum_{h \in [N_h]} \left( \mathbf{X}_{\mathbf{q}-s(h),:} \mathbf{W}^{(h)}_{\text{val}} \mathbf{W}^{(h)}_{\text{out}} \right)  + \mathbf{b}_{\text{out}} \\
    &= \sum_{h \in [N_h]} \left( \mathbf{X}_{\mathbf{q}-s(h),:} \mathbf{W}^{(h)} \right)  + \mathbf{b}_{\text{out}} \\
    &= \sum_{\kappa \in \nabla_{k}} \mathbf{X}_{\mathbf{q}-\kappa,:} \mathbf{W}^{\text{(conv)}}_{\kappa,:,:} + \mathbf{b}_{\text{out}} \\
    &= \text{Conv2D} \left( \mathbf{X}; \mathbf{W}^{\text{(conv)}} \right)_{\mathbf{q}, :} + \mathbf{b}_{\text{out}},
\end{align}
where $\mathbf{W}^{\text{(conv)}}_{s(h)} \coloneqq \mathbf{W}^{(h)} \in \mathbb{R}^{D_{\text{emb}} \times D_{\text{emb}}}$ is a weight matrix of 2-dimensional convolutional kernel, $\mathbf{W}^{\text{(conv)}} \in \mathbb{R}^{k \times k \times D_{\text{emb}} \times D_{\text{out}}}$, at position $s(h)$\footnote{This proof is primarily based on the work of Cordonnier \etal~\cite{cordonnier2021ontherelationship}.}. \QEDA

\begin{figure*}[t]
    \begin{center}

    \scalebox{0.43}{
    \centering
        \includegraphics{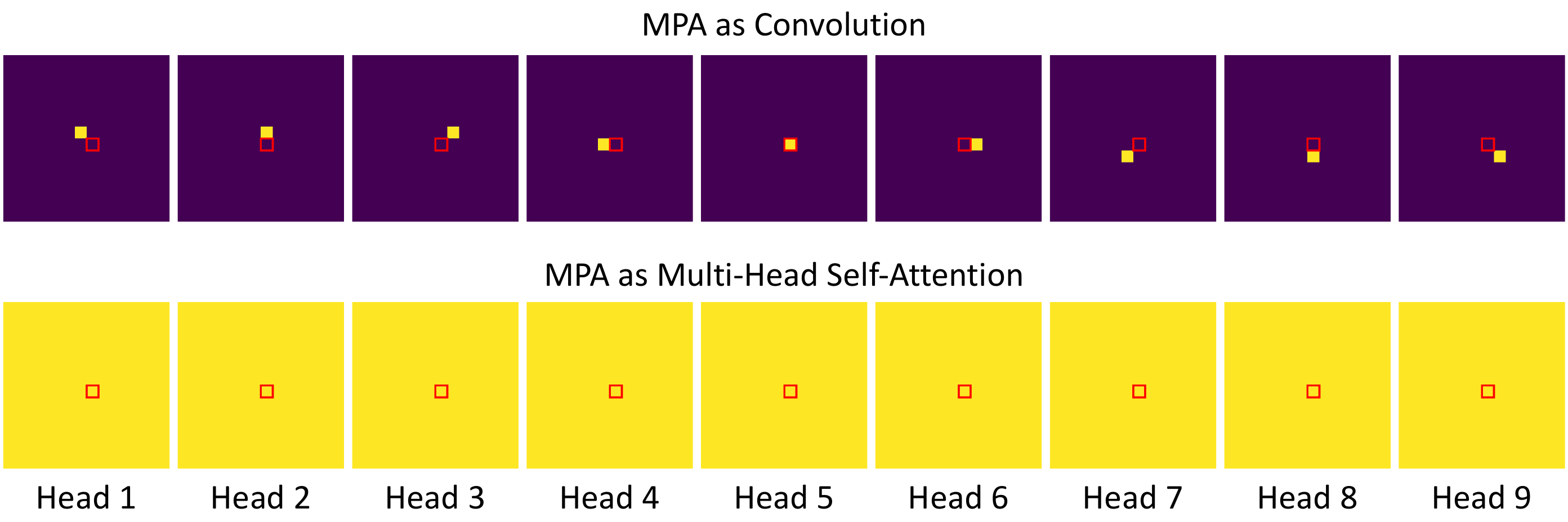}
    }
    \vspace{-0.2mm}
    \caption{MPA layers with $N_h = 9$ heads in two extreme cases of semi-dynamic attention (top) and position bias injection (bottom) which respectively express $3 \times 3$ convolution and MHSA.} 
    \vspace{-6.0mm}
    \label{fig:supp_conv_sa}
    \end{center}   
\end{figure*}

\smallbreak\noindent\textbf{MPA as an multi-head self-attention.} 
Assume that the content-based attention is scaled dot product between queries and keys, \ie, $\Phi^{(h)}_{\text{c}}(\mathbf{X}) = \exp(\tau \mathbf{Q}^{(h)}, \mathbf{K}^{(h)\top})$, and the bias term of the second instance normalization in $\Phi^{(h)}_{\text{p}}$ is set to some large number such that $\beta_{\text{p2}}^{(h)} = \infty$ for all heads $h$, \ie, {\em an extreme (global) position bias injection}.
Note that the latter assumption gives $\Phi^{(h)}_{\text{p}}(\mathbf{R})_{\mathbf{q}, \mathbf{k}} = 1$ for all $\mathbf{q}, \mathbf{k} \in \mathcal{P}$ and $h \in [N_h]$ according to Eq.~\ref{eq:beta_infty}.
Now consider Peripheral-Attention:
\begin{align}
    \text{Peripheral-Attention}^{(h)}(\mathbf{X}, \mathbf{R}) &= \text{Normalize}\left[\Phi^{(h)}_{\text{c}}(\mathbf{X}) \odot \Phi^{(h)}_{\text{p}}( \mathbf{R} ) \right] \mathbf{V}^{(h)} \\
    &= \text{Normalize}\left[ \exp(\tau \mathbf{Q}^{(h)}, \mathbf{K}^{(h)\top}) \odot J_{HW, HW} \right] \mathbf{V}^{(h)} \\ 
    &= \text{softmax}\left( \tau \mathbf{Q}^{(h)}, \mathbf{K}^{(h)\top} \right) \mathbf{V}^{(h)} \\
    &= \text{Self-Attention}^{(h)}(\mathbf{X}),
\end{align}
where $J_{HW, HW} \in \mathbb{R}^{HW \times HW}$ refers to all-one matrix. \QEDA

Figure~\ref{fig:supp_conv_sa} illustrates MPA layers in two extreme cases of semi-dynamic attention and position bias injection with $N_h = k^2 = 9$.

\clearpage


\clearpage

\section{Additional Results and Analyses}
\label{sec:supp_results_analyses}
In this section, we provide additional results and analyses of the proposed method.

\smallbreak\noindent\textbf{Nonlocality and impact measure comparisons with other baseline methods.}
To investigate how the peripheral position encoding $\Phi_{\text{p}}$ benefits PerViT, we compare the nonlocality measure of PerViT-T with that of our baseline DeiT-T~\cite{touvron2021deit} and DeiT-T with state-of-the-art RPE method, \ie, iRPE-K~\cite{wu2021irpe}.
To measure the nonlocality $\Omega_{\text{a}}$ (Eq.~\ref{eq:nonlocality} of our main paper) of the other models, we use $\Phi_{\text{a}} = \exp(\mathbf{Q}\mathbf{K}^{\top}) \in \mathbb{R}^{HW \times HW}$ for DeiT, and $\Phi_{\text{a}} = \exp(\mathbf{Q}\mathbf{K}^{\top} + \mathbf{R}_{\text{rpe-K}}) \in \mathbb{R}^{HW \times HW}$ for iRPE\footnote{The input to the exponential function is normalized to prevent large activations following the implementation of softmax in PyTorch~\cite{pytorch}. For iRPE, we use contextual product method proposed in~\cite{wu2021irpe}.}.

\begin{figure*}[h]
    \begin{center}

    \scalebox{0.33}{
    \centering
        \includegraphics{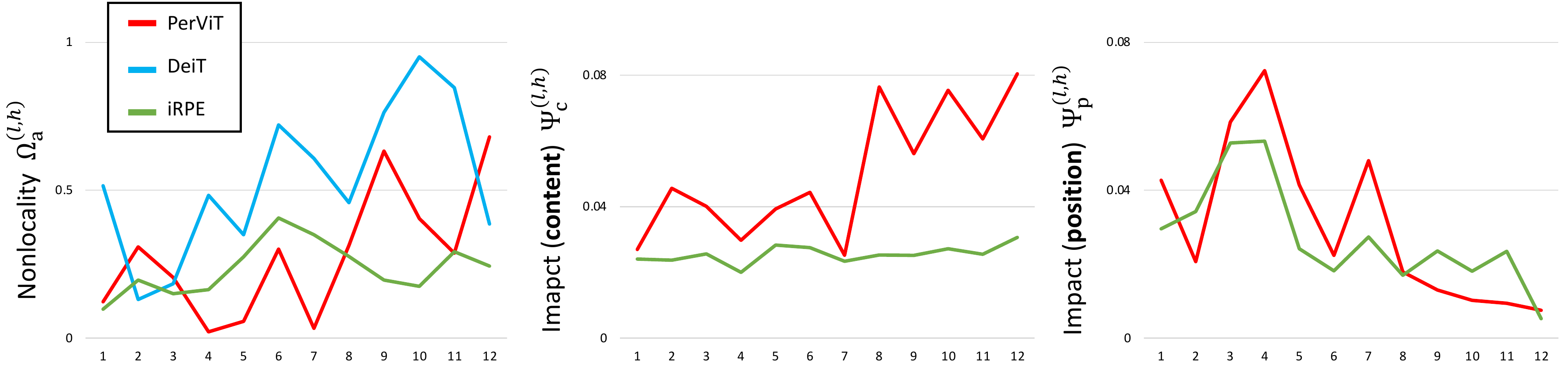}
    }
    \vspace{-0.5mm}
    \caption{Nonlocality (left) and impact measure (middle: content-based $\Psi_{\text{c}}$, right: position-based $\Psi_{\text{p}}$) comparisons between DeiT~\cite{touvron2021deit}, iRPE~\cite{wu2021irpe}, and PerViT (ours).}
    \vspace{-3.0mm}
    \label{fig:nonlocality_deit_irpe}
    \end{center}   
\end{figure*}

As seen in the left of Fig.~\ref{fig:nonlocality_deit_irpe},  DeiT learns to attend more locally in early layers compared to the late ones but its overall nonlocality is higher than that of PerViT due to the absence of position information $\Phi_{\text{p}}$, implying that position encoding effectively encourages higher locality for fine-grained pattern recognition.
Meanwhile, the iRPE effectively imposes locality on attentions but it provides similar magnitudes of nonlocality across different layers, \ie, local attention at every layer, while those of PerViT and DeiT highly varies from early (local) to late (global) layers.

\begin{figure*}[h]
    \begin{center}

    \scalebox{0.35}{
    \centering
        \includegraphics{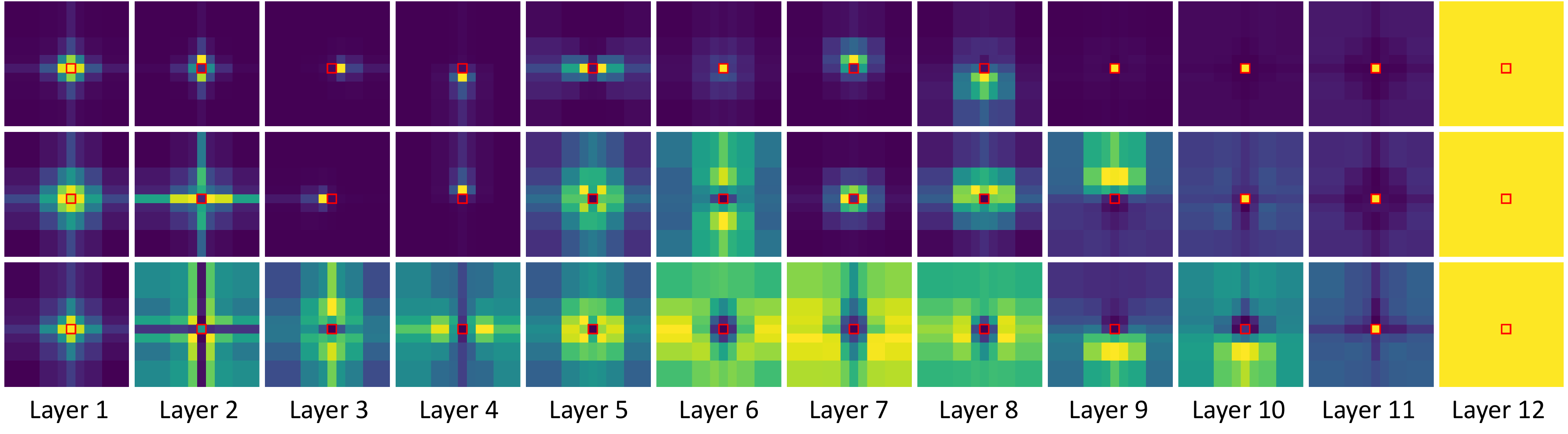}
    }
    \vspace{-0.5mm}
    \caption{Learned position-based attention $\Phi_{\text{p}}$ of iRPE-K product method~\cite{wu2021irpe}.}
    \vspace{-6.0mm}
    \label{fig:attention_irpe}
    \end{center}   
\end{figure*}

\begin{figure*}[h]
    \begin{center}

    \scalebox{0.37}{
    \centering
        \includegraphics{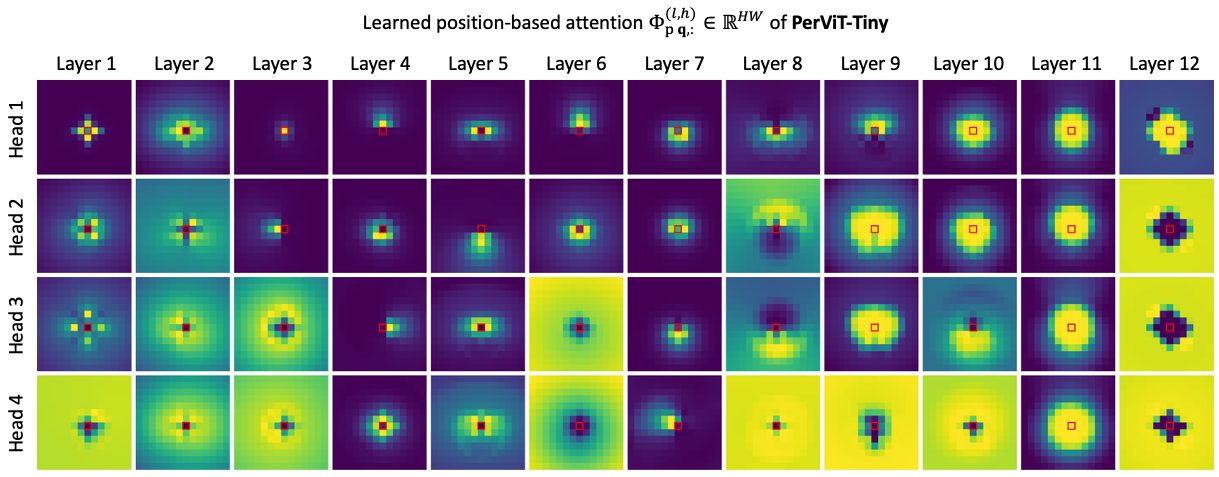}
    }
    \vspace{-0.5mm}
    \caption{Learned position-based attentions $\Phi_{\text{p} \ \mathbf{q},:}^{(l,h)}$ of PerViT-Tiny.} 
    \vspace{-6.0mm}
    \label{fig:phip_tiny}
    \end{center}   
\end{figure*}

To demonstrate this phenomenon, we plot and compare the impacts of $\Phi_{\text{c}}$ and $\Phi_{\text{p}}$\footnote{We separate $\Phi_{\text{a}} = \exp(\mathbf{Q}\mathbf{K}^{\top} + \mathbf{R}_{\text{rpe-K}}) = \exp(\mathbf{Q}\mathbf{K}^{\top}) \ \odot \ \exp(\mathbf{R}_{\text{rpe-K}})$ in iRPE into two terms of content- and position-based attentions: $\Phi_{\text{c}} = \exp(\mathbf{Q}\mathbf{K}^{\top})$ and $\Phi_{\text{p}} = \exp(\mathbf{R}_{\text{rpe-K}})$.} on $\Phi_{\text{a}}$ in the middle and right of Fig.~\ref{fig:nonlocality_deit_irpe} and visualize learned position-based attention $\Phi_{\text{p}}$ of iRPE in Fig.~\ref{fig:attention_irpe}.
From the plots, we note that the impacts of content- and position-based attentions in iRPE also have relatively low variance compared to PerViT.
We conjecture that RPE transformation layer in~\cite{wu2021irpe} only consists of a single linear projection while being shared across every layer in the network, thus providing attention maps with less diversity in terms of both shapes and sizes (Fig.~\ref{fig:attention_irpe} {\em vs.} Fig.~\ref{fig:phip_tiny}).

Interestingly, we also observe that attended regions in iRPE complement each other to cover the whole visual field, quite similarly to PerViT.
For example, the attended regions in the first and second heads of Layer 3 form horizontal attention in the {\em central region} while the third head attends to the rest in the {\em peripheral region}.
Note that most layers (Layer 2-10) behave similarly as in {\em human peripheral vision}, which again support the main motivation of this work.

From this study, we draw following conclusions: 
(i) Sufficiently high locality imposed by position bias improves vision transformer.
(ii) The position-based attentions should be in diverse shapes and sizes across different layers.
(iii) Without any special supervisions, RPEs learn to model {\em peripheral vision} solely based on training images, proving that our work poses a promising direction.

\clearpage

\begin{figure*}[h]
    \begin{center}

    \scalebox{0.30}{
    \centering
        \includegraphics{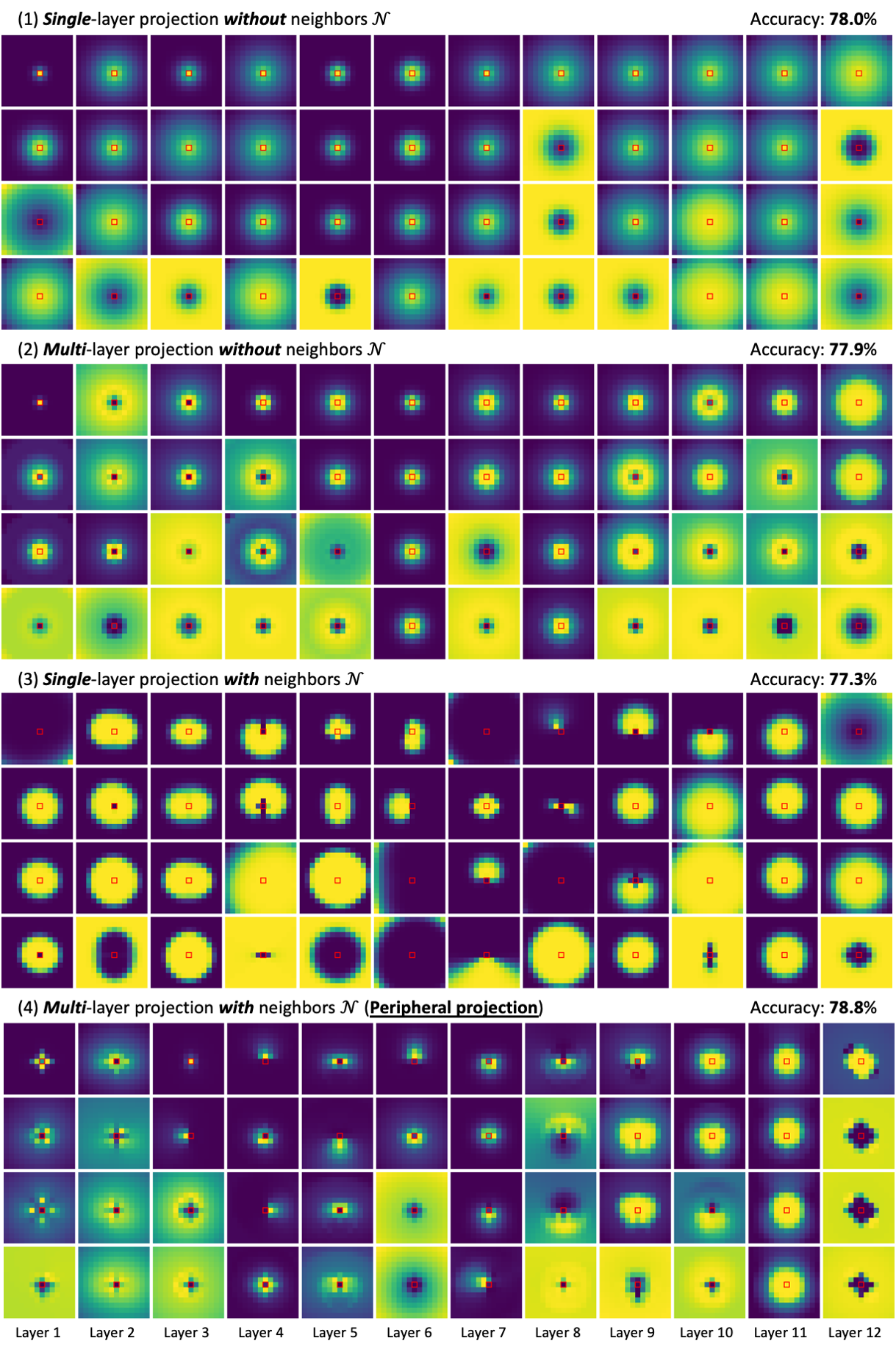}
    }
    \vspace{-0.5mm}
    \caption{Learned position-based attention $\Phi_{\text{p}}$ with different network design choices.}
    \vspace{-4.0mm}
    \label{fig:peripheral_proj_abl}
    \end{center}   
\end{figure*}

\smallbreak\noindent\textbf{Qualitative comparison between different network designs of} $\Phi_{\text{p}}$\textbf{.}
In Tab.~\ref{table:phip_study} of the main paper, we explored different network designs of $\Phi_{\text{p}}$ in PerViT-T: Euc (single-layer proj. w/o $\mathcal{N}$), Euc+ML (multi-layer proj. w/o $\mathcal{N}$), and Euc+$\mathcal{N}$+ML (multi-layer proj. with $\mathcal{N}$, \ie, \textbf{ours}).
We visualize their learned position-based attentions $\Phi_{\text{p}}$ in Fig.~\ref{fig:peripheral_proj_abl}.
As discussed in Sec.~\ref{sec:mpa_main} and can be seen from first and third group of attentions, the single-layer projections are only able to provide Gaussian-like attention maps.
We observe that referring neighbors $\mathcal{N}$ with single-layer severely damages performance;
we suspect that unnecessarily large attention scores at distant positions (possibly caused by neighborhood aggregation) hinder the ability to focus on local patterns as seen from nonlocal circular attentions in the third group.
The multi-layer (ML) projection in the second group helps the model in forming (torus-shaped) peripheral regions but it still poses undesirable rotational symmetric property in feature transformations.
Note that the proposed ML projection with $\mathcal{N}$ (the last group of attentions) gently breaks the rotational symmetric property while retaining the ability to form peripheral regions to a sufficient extent with a significant performance boost, implying that modelling effective attention-based peripheral vision demands both designs of multi-layer and neighborhood.
\clearpage
\smallbreak\noindent\textbf{Further analyses on nonlocality} $\Omega_{\text{a}}$ \textbf{and} $\Omega_{\text{p}}$\textbf{.}
In Tab.~\ref{table:ablation_main} of the main paper, we have analyzed how the absence of each component in PerViT ($\Phi_{\text{p}}$, $\Phi_{\text{c}}$, C-stem, and CPE) impacts on the performance, \eg, ImageNet accuracy.
In this study, we investigate their impact on {\em both performance and locality} in order to reveal their relationship in vision transformers.

\begin{figure*}[h]
    \begin{center}

    \scalebox{0.3}{
    \centering
        \includegraphics{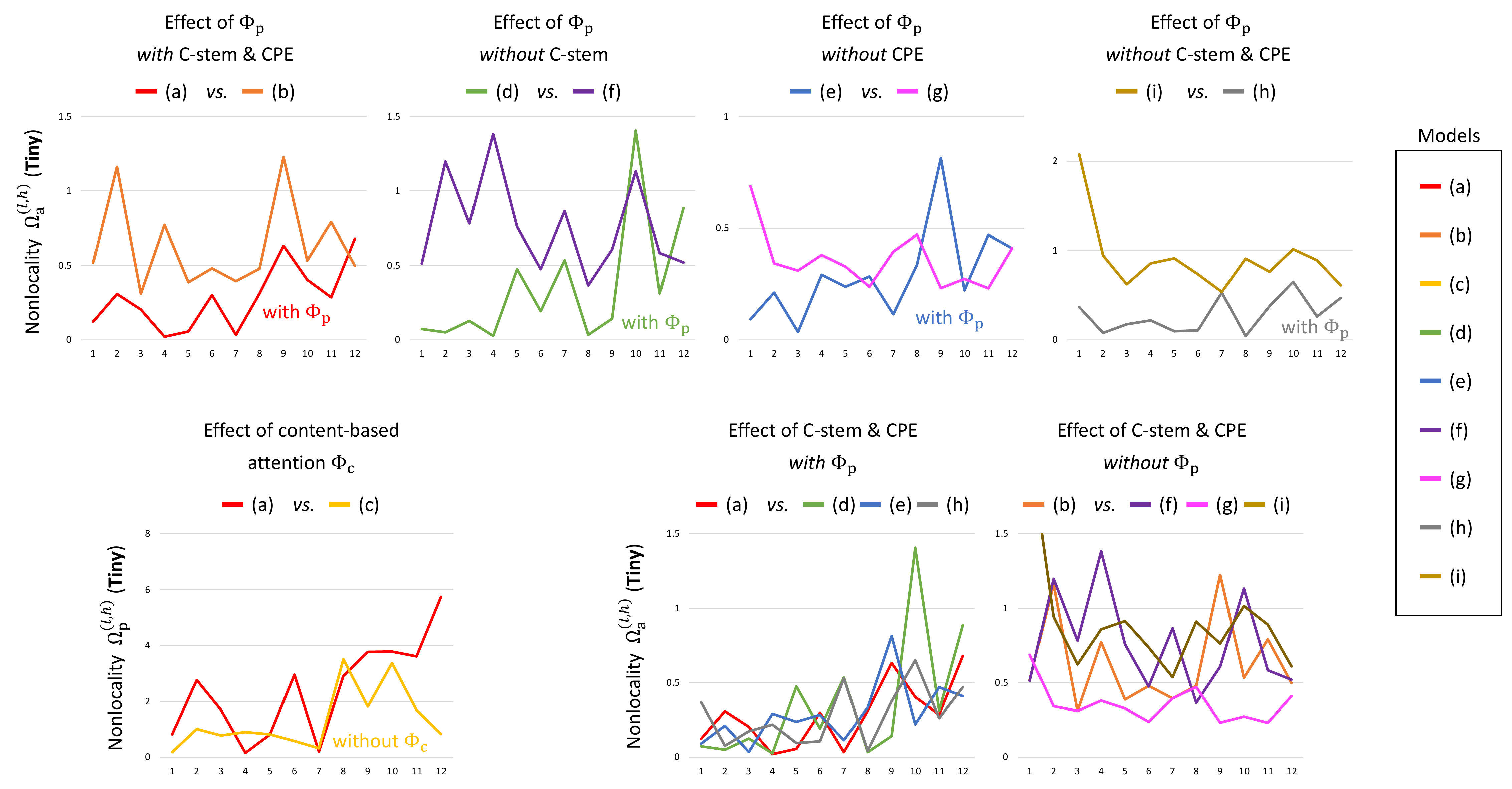}
    }
    \vspace{-0.5mm}
    \caption{The effect of each component in PerViT to nonlocality measures $\Omega_{\text{a}}$ (top 4 and bottom-right 2 plots) and $\Omega_{\text{p}}$ (bottom-left plot).} 
    \vspace{-6.0mm}
    \label{fig:nonlocality_inspection}
    \end{center}   
\end{figure*}

As seen from Tab.~\ref{table:ablation_supp}\footnote{We copy the results of Tab.~\ref{table:ablation_main} in the main paper for the ease of view.}, there exist noticeable performance gaps between the models (b, f, g, h) (without $\Phi_{\text{p}}$) and (a, d, e, i) (with $\Phi_{\text{p}}$).
To see if the improvements are from the locality imposed by $\Phi_{\text{p}}$, we plot and compare the nonlocality measures of the models in Fig.~\ref{fig:nonlocality_inspection}.
As seen from the top four plots, $\Phi_{\text{p}}$ always imposes early local attentions without any exceptions.
The results prove the necessity of {\em local transformations in the early stages} and again support recent research directions towards augmenting early convolutions in vision transformer architectures~\cite{dai2021coatnet, li2022uniformer, ramachandran2019sasa, srinivas2021botnet, xiao2021early}.

The bottom-left plot of Fig.~\ref{fig:nonlocality_inspection} reveals the impact of content-based attention $\Phi_{\text{c}}$ on attention locality.
Without adaptive attention (model (c)), $\Phi_{\text{p}}$ imposes stronger locality on every layer.
Compared to (a) (Fig.~\ref{fig:phip_tiny}), (c) performs `hard' attentions as seen from Fig.~\ref{fig:phip_wo_phic}; 
especially, the early attentions look noticeably similar to convolutions as illustrated at the top of Fig.~\ref{fig:supp_conv_sa}.
Removing dynamicity encourages the model (c) to be more `convolutional' as it loses the ability to adaptively collect relevant features over broader areas, exploiting only a few relevant ones in local regions statically.

Two plots at the bottom-right of Fig.~\ref{fig:nonlocality_inspection} show the impact of convnets, \eg, C-stem and CPE, on locality of mixed attention $\Phi_{\text{a}}$.
We find that the absence of the convnets hardly affects the locality.
In presence of $\Phi_{\text{p}}$, model performs early local and late global attention like PerViT (a).
In absence of $\Phi_{\text{p}}$, the overall nonlocality increases for all models (b, f, g, i).
The results reveal that built-in locality of convolutional layers does not have direct effects on locality of the self-attention layers.

\begin{table}[h]
	\begin{minipage}{0.33\linewidth}
		\centering
        \vspace{-3.0mm}
        \caption{Ablation study.}
        \vspace{2.0mm}
        \label{table:ablation_supp}
        \scalebox{0.65}{
                    \begin{tabular}{cccccc}
                    
                    \toprule
                    
                    Ref. & $\Phi_{\text{p}}$ & $\Phi_{\text{c}}$ & C-stem & CPE & acc. \\
                    
                    \midrule
                    
                    (a) & \cmark  & \cmark  & \cmark  & \cmark & 78.8 \\
                    (b) & \xmark  & \cmark  & \cmark  & \cmark & 77.3 \\
                    (c) & \cmark  & \xmark  & \cmark  & \cmark & 76.8 \\
                    (d) & \cmark  & \cmark  & \xmark  & \cmark & 77.8 \\
                    (e) & \cmark  & \cmark  & \cmark  & \xmark & 78.1 \\
                    (f) & \xmark  & \cmark  & \xmark  & \cmark & 76.3 \\
                    (g) & \xmark  & \cmark  & \cmark  & \xmark & 76.7 \\
                    (h) & \cmark  & \cmark  & \xmark  & \xmark & 76.5 \\
                    (i) & \xmark  & \cmark  & \xmark  & \xmark & 72.3 \\
                    
                    \bottomrule
                    
                    \end{tabular}
                }
        \vspace{0.0mm}
	\end{minipage}\hfill
	\begin{minipage}{0.65\linewidth}
        \begin{center}
    
        \scalebox{0.25}{
        \centering
            \includegraphics{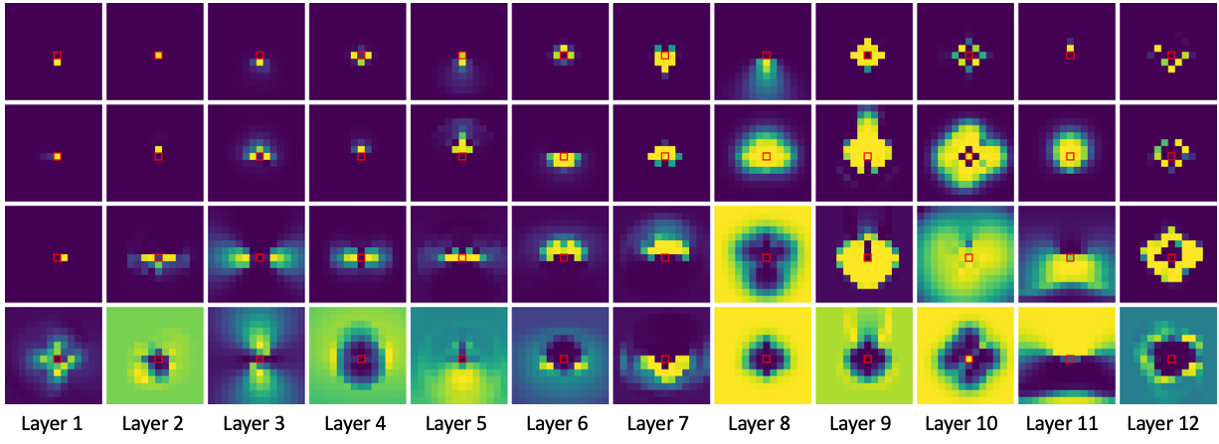}
        }
        \vspace{-3.0mm}
        \captionof{figure}{Learned position-based attention $\Phi_{\text{p}}$ of model (c).} 
        \vspace{-0.0mm}
        \label{fig:phip_wo_phic}
        \end{center}   
	\end{minipage}
	\vspace{-3.0mm}
\end{table}

\clearpage

\begin{figure*}[h]
    \begin{center}

    \scalebox{0.37}{
    \centering
        \includegraphics{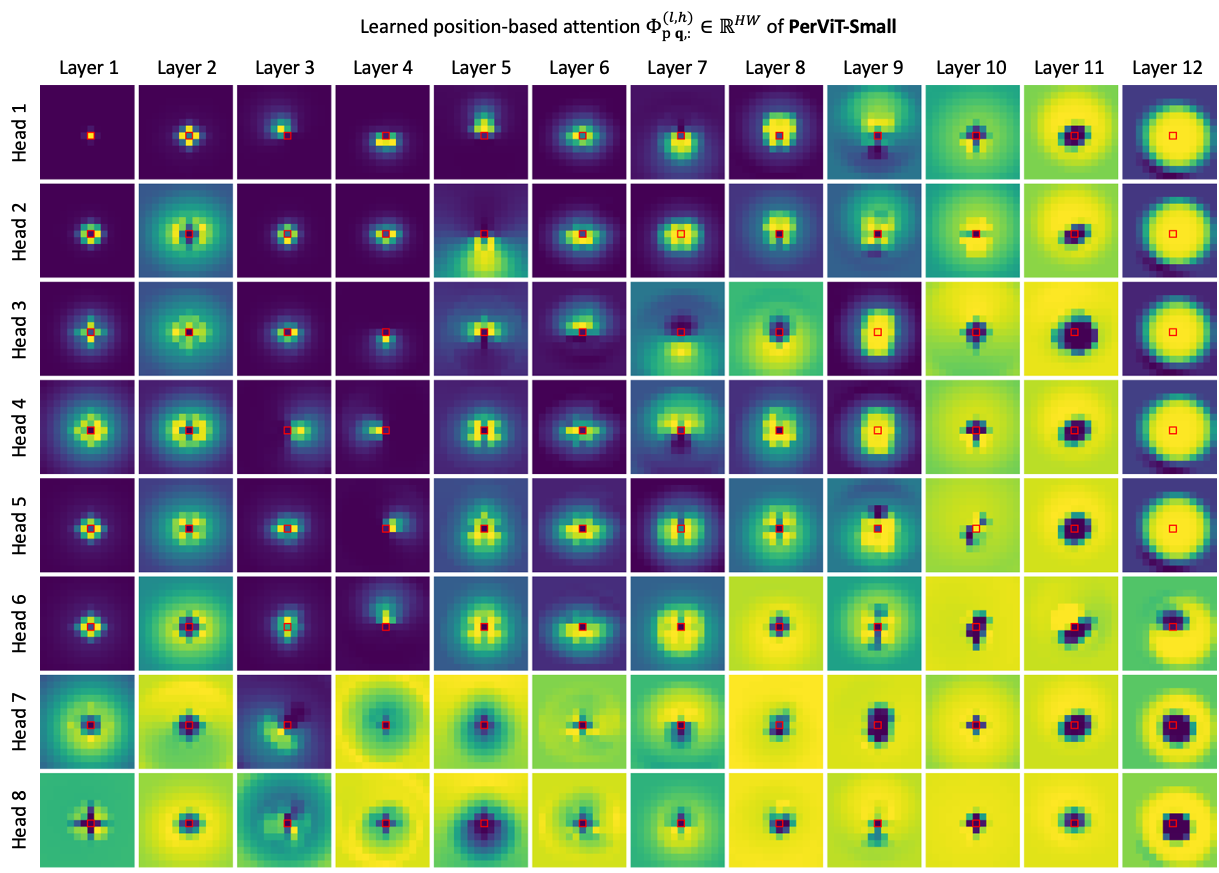}
    }
    \vspace{-0.5mm}
    \caption{Learned position-based attentions $\Phi_{\text{p} \ \mathbf{q},:}^{(l,h)}$ of PerViT-Small.} 
    \vspace{-6.0mm}
    \label{fig:phip_small}
    \end{center}   
\end{figure*}

\begin{figure*}[h]
    \begin{center}

    \scalebox{0.37}{
    \centering
        \includegraphics{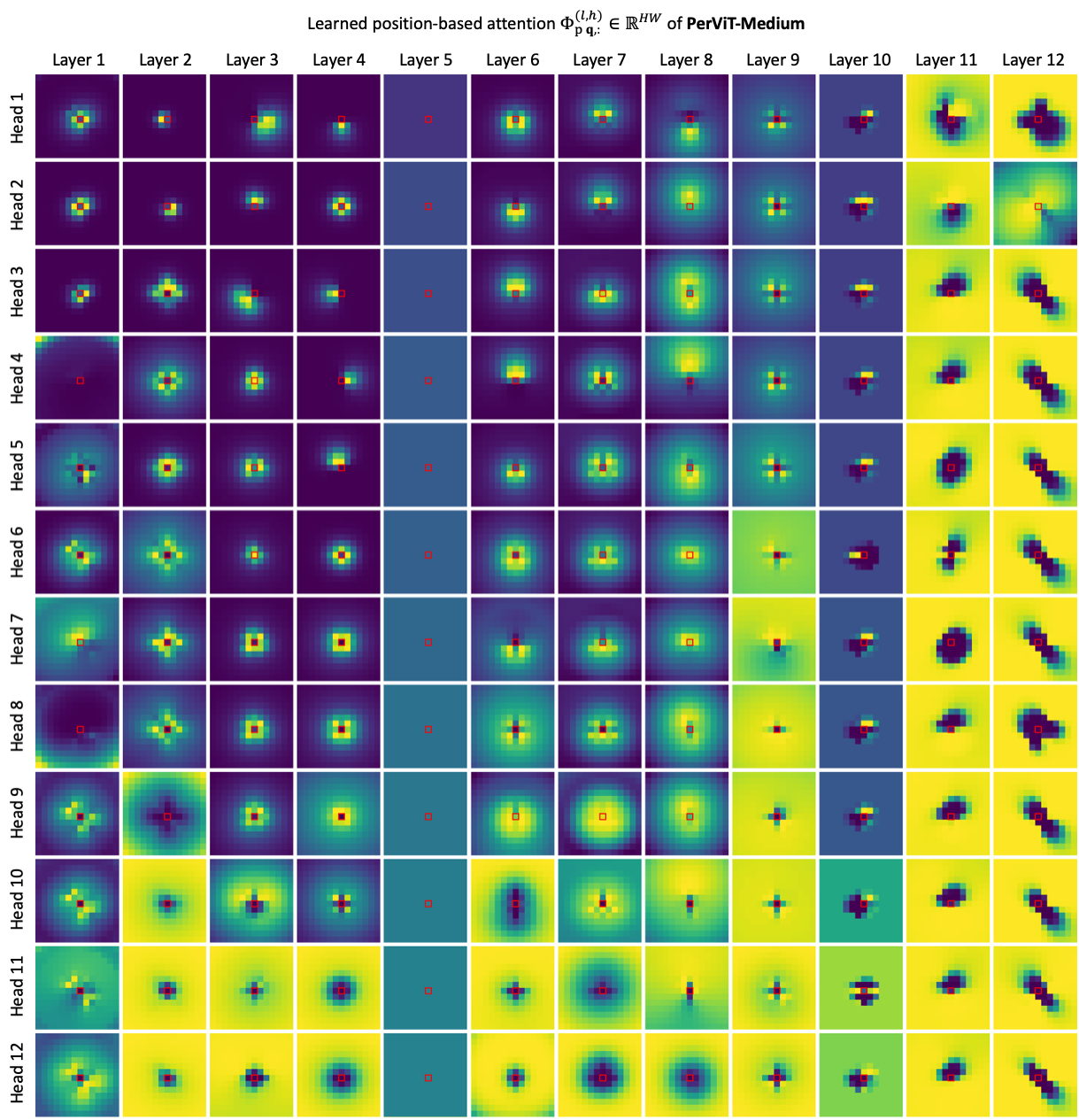}
    }
    \vspace{-0.5mm}
    \caption{Learned position-based attentions $\Phi_{\text{p} \ \mathbf{q},:}^{(l,h)}$ of PerViT-Medium.} 
    \vspace{-6.0mm}
    \label{fig:phip_medium}
    \end{center}   
\end{figure*}

\begin{figure*}[h]
    \begin{center}

    \scalebox{0.25}{
    \centering
        \includegraphics{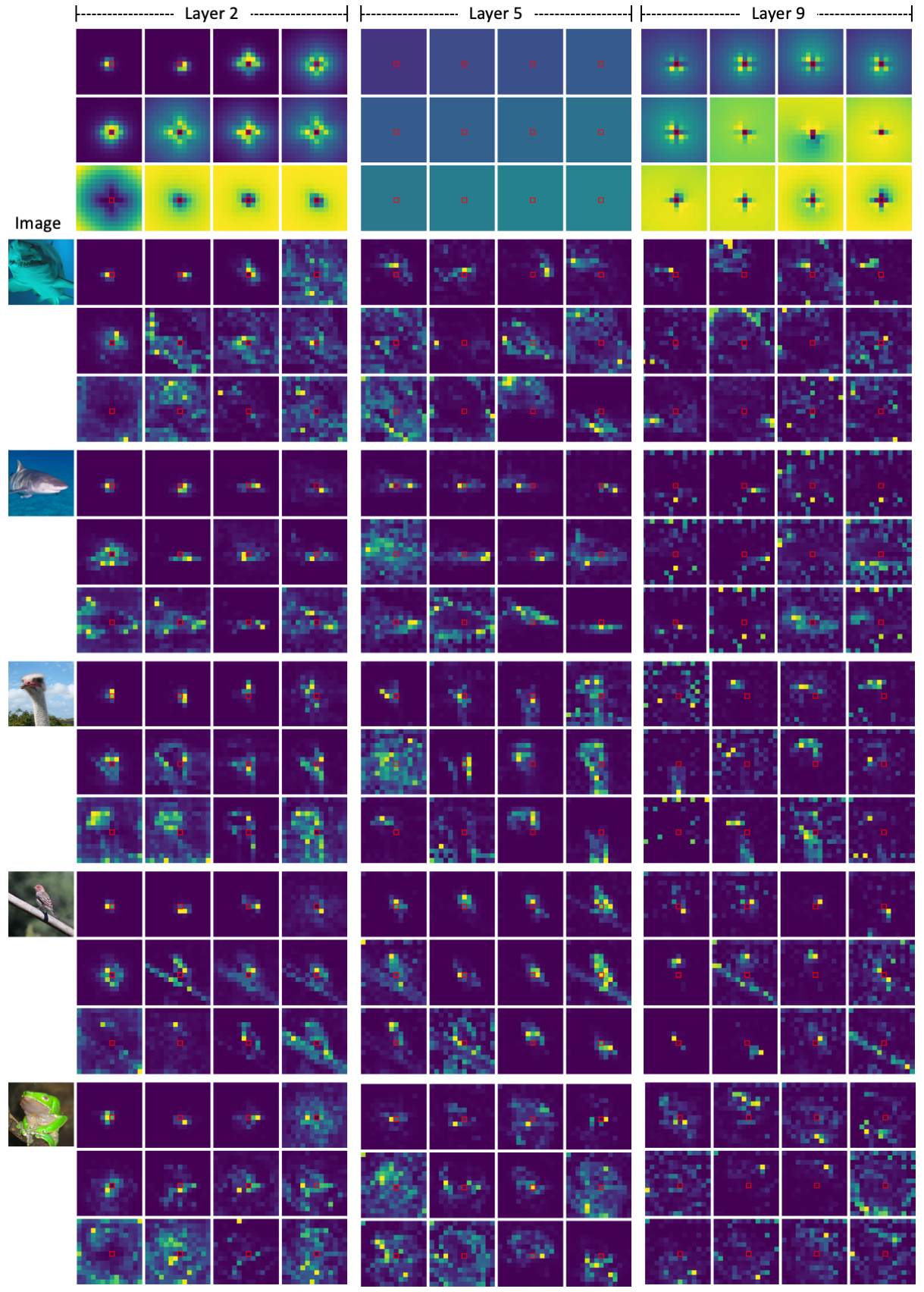}
    }
    \vspace{-4.0mm}
    \caption{Visualization of attentions $\Phi_{\text{p}}^{(l,h)}$ and $\Phi_{\text{a}}^{(l,h)}$ of PerViT-Medium.} 
    \vspace{-6.0mm}
    \label{fig:qual_medium}
    \end{center}   
\end{figure*}

\smallbreak\noindent\textbf{Visualization of learned position-based attention $\Phi_{\text{p}}$.} 
Figures~\ref{fig:phip_tiny}, \ref{fig:phip_small}, and \ref{fig:phip_medium} depict learned position-based attentions $\Phi_{\text{p}}$ in PerViT-Tiny, Small, and Medium respectively where the attention maps at each layer are sorted in the order of nonlocality measure.
We observe that the overall nonlocality of a model, \eg, brightness of the attention maps, noticeably increases as the model size grows as discussed in Sec.~\ref{sec:innerworking_main} of the main paper.
Interestingly, for all three models, the query position $\mathbf{q}$, \ie, the center, is not attended in the most of the learned attention; we hypothesize that a sufficient amount of transformations on query position is already done by $3 \times 3$ depth-wise convolutions in CPE and point-wise ($1 \times 1$) convolutions in MLP layers so query position no longer require further transformations in MPA which thereupon focuses mostly on {\em peripheral regions}.
We also observe that MPA at Layer 5 of the Medium model performs uniform, global attentions at different scales just like MHSA as seen in Fig.~\ref{fig:qual_medium}.

\clearpage

\clearpage

\begin{figure*}[h]
    \begin{center}

    \scalebox{0.25}{
    \centering
        \includegraphics{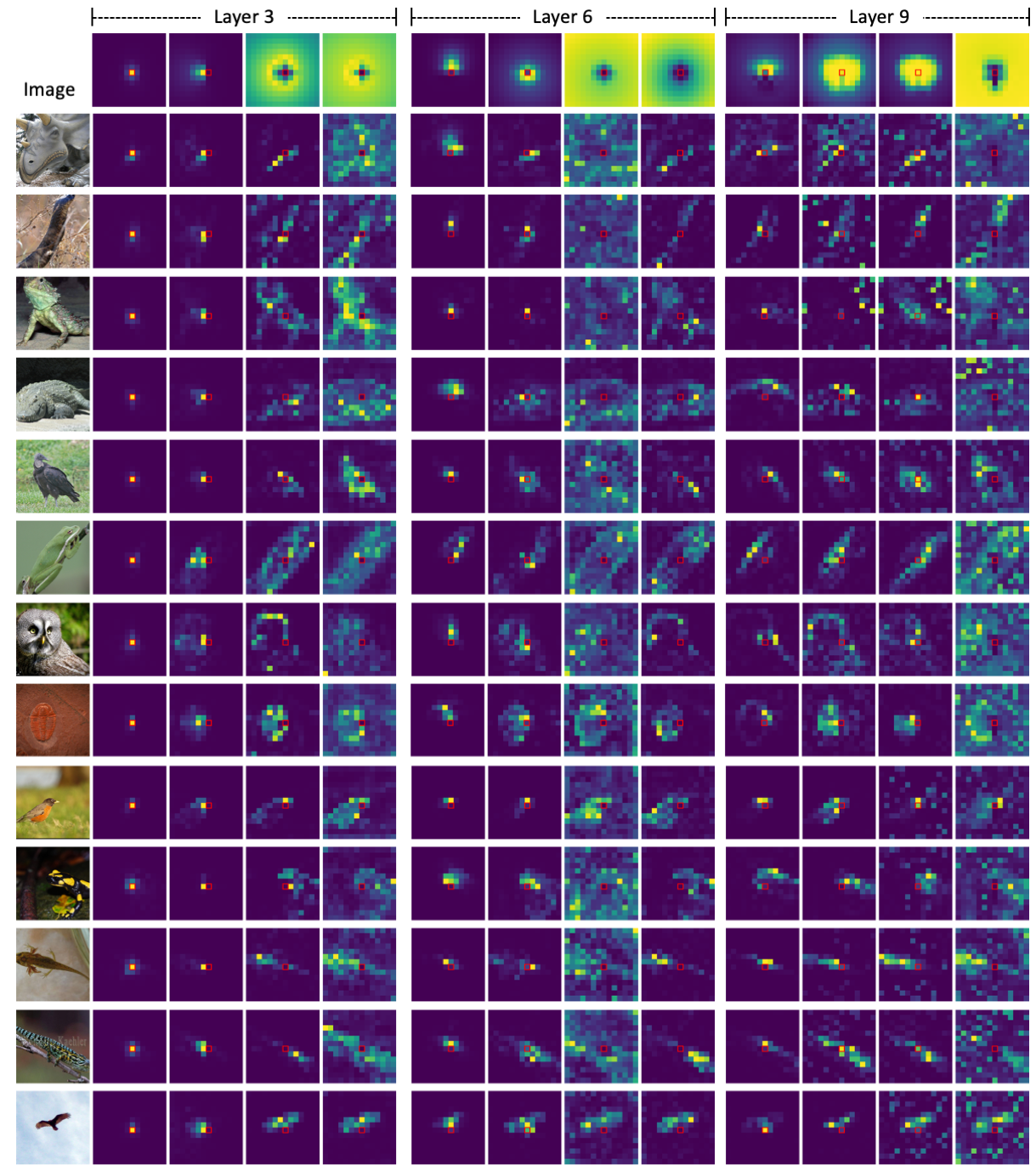}
    }
    \vspace{-4.0mm}
    \caption{Visualization of attentions $\Phi_{\text{p}}^{(l,h)}$ and $\Phi_{\text{a}}^{(l,h)}$ of PerViT-Tiny.} 
    \vspace{-6.0mm}
    \label{fig:qual_tiny}
    \end{center}   
\end{figure*}

\begin{figure*}[h]
    \begin{center}

    \scalebox{0.25}{
    \centering
        \includegraphics{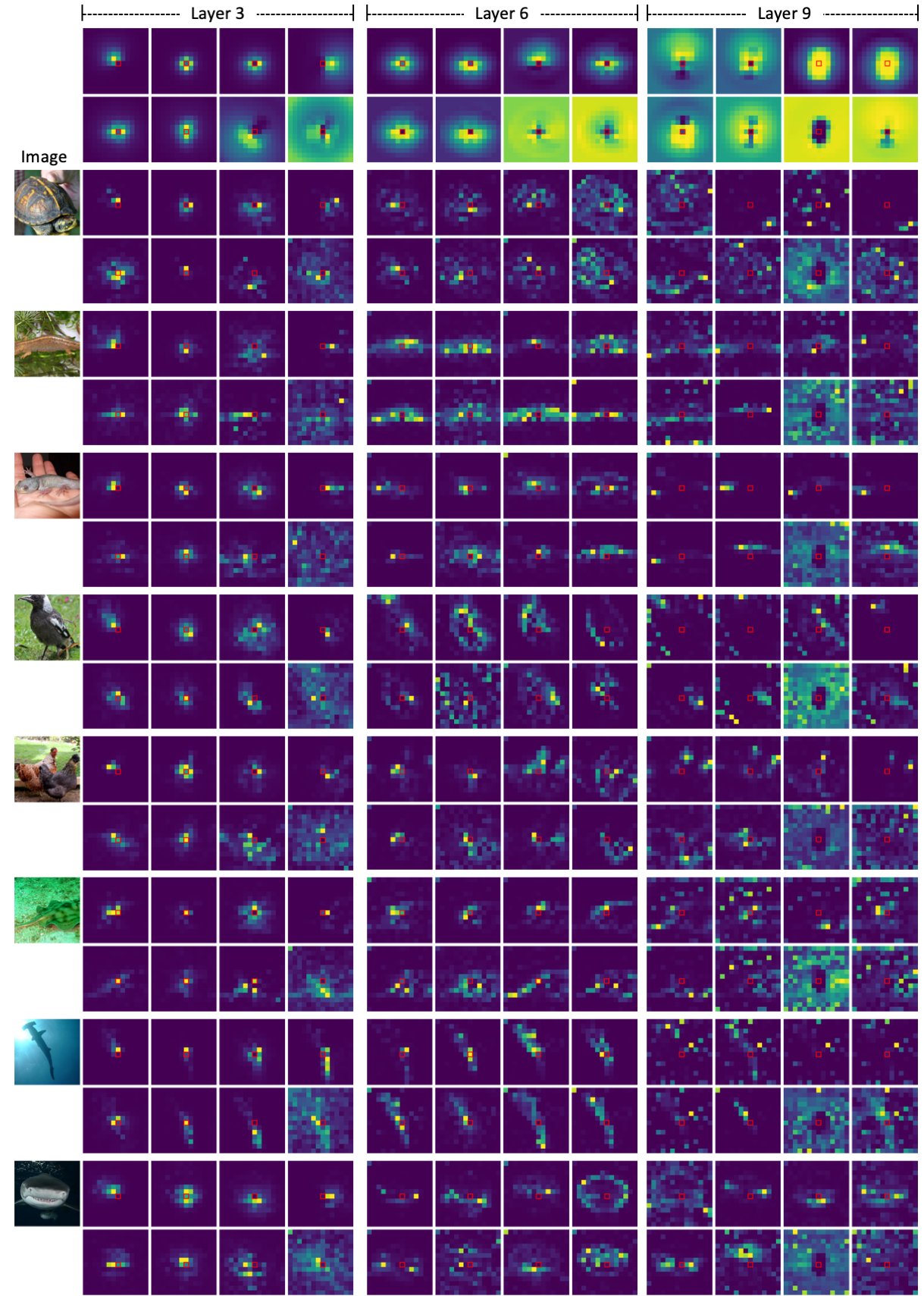}
    }
    \vspace{-4.0mm}
    \caption{Visualization of attentions $\Phi_{\text{p}}^{(l,h)}$ and $\Phi_{\text{a}}^{(l,h)}$ of PerViT-Small.} 
    \vspace{-6.0mm}
    \label{fig:qual_small}
    \end{center}   
\end{figure*}

\smallbreak\noindent\textbf{Further analyses on qualitative results.} 
We visualize sample attention maps of PerViT Tiny, Small, and Medium in Figs.~\ref{fig:qual_tiny}, \ref{fig:qual_small}, and \ref{fig:qual_medium} respectively where position-based attentions $\Phi_{\text{p}}$ are shown at the top, and sample images with their mixed attentions $\Phi_{\text{a}}$ are listed below.
We pick three layers among early to late layers to investigate how MPAs form mixed attention across different layers.

At early levels, \eg, Layers 2-3, about half the number of heads in each MPA forms (semi-) static/local attentions in central regions while the others provide relatively dynamic/global attentions in the peripheral regions, all of which complement each other to cover whole visual field.
Note that the attention scores in early to intermediate levels, \eg, Layers 2-6, mostly fall inside the object of interest, capturing relevant visual patterns for effective image classification.
However, at later level, \eg, Layer 9, position-based attention scores $\Phi_{\text{p}}$ are formed in mid- and far-peripheral regions, and mixed attention scores $\Phi_{\text{a}}$ are scattered over the whole spatial area.
We conjecture that the late layer MPAs collect complementary features which the network missed along early to intermediate feature processing pathways so as to reinforce the image embeddings for the final prediction.


\clearpage

\begin{figure*}[h]
    \begin{center}

    \scalebox{0.36}{
    \centering
        \includegraphics{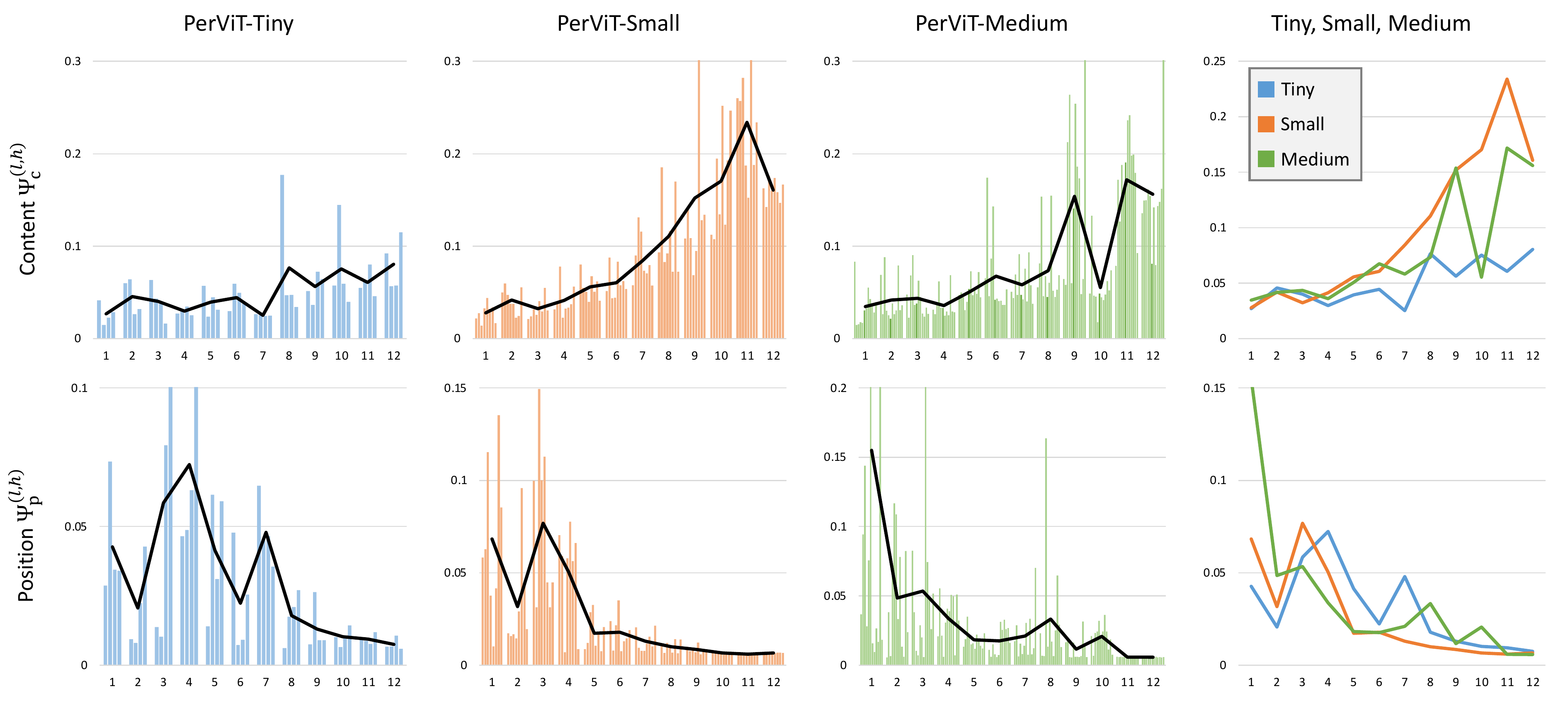}
    }
    \vspace{-0.5mm}
    \caption{The measure of impact (x-axis: layer index, y-axis: the impact metric $\Psi_{*}$). Each bar graph shows the measure of a single head (\# heads ($N_h$) are set to 4, 8, and 12 for Tiny, Small, and Medium models respectively), and the solid lines represent the trendlines which follow the average values of layers.}
    \vspace{-0.0mm}
    \label{fig:impact_measure}
    \end{center}   
\end{figure*}

\begin{figure*}[h]
    \begin{center}

    \scalebox{0.36}{
    \centering
        \includegraphics{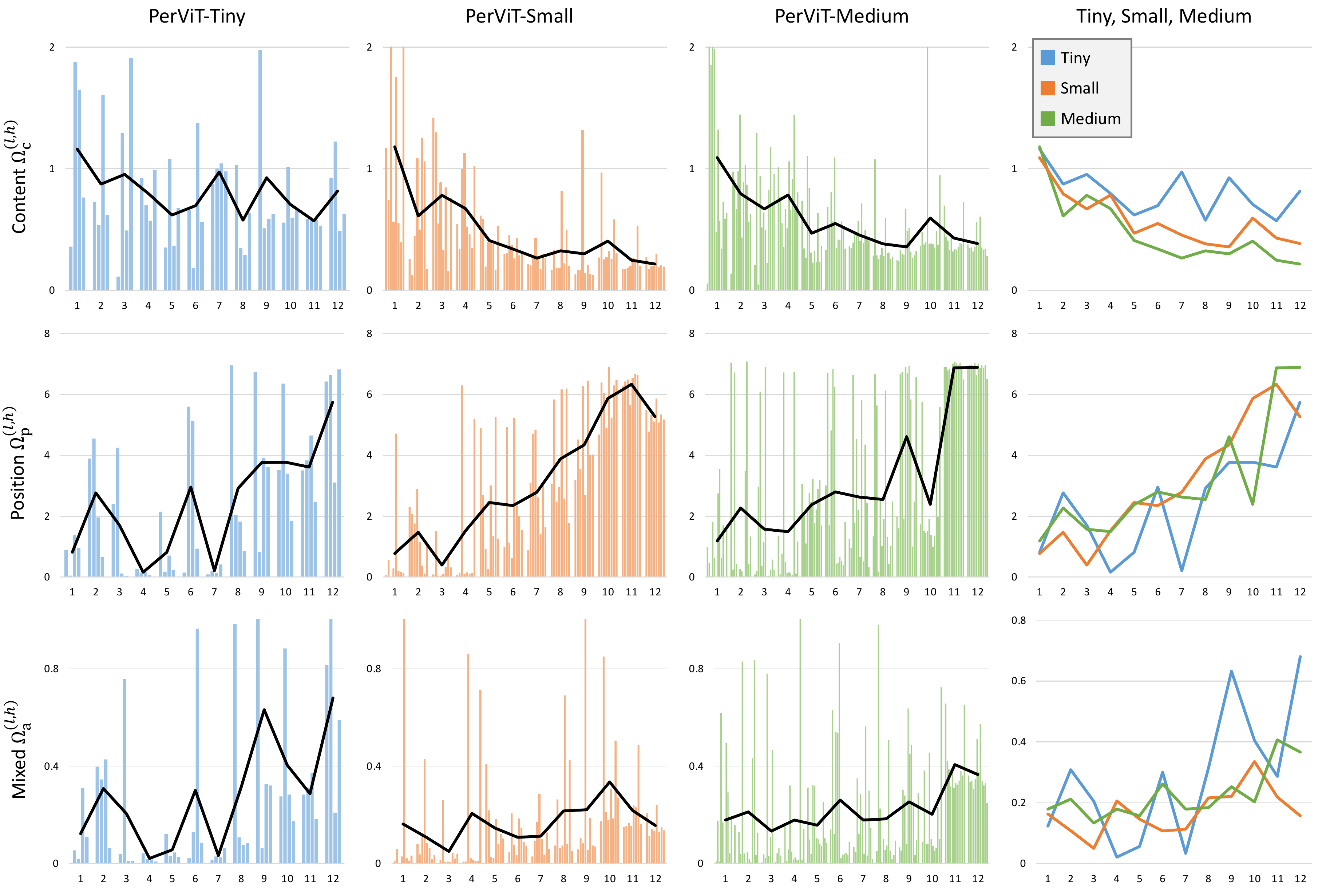}
    }
    \vspace{-0.5mm}
    \caption{The measure of nonlocality (x-axis: layer index, y-axis: the nonlocality measure $\Omega{*}$).} 
    \vspace{-6.0mm}
    \label{fig:nonlocality_measure}
    \end{center}   
\end{figure*}

\smallbreak\noindent\textbf{The impact and nonlocality measures of} $\Phi_{\text{p}}^{(l,h)}$ \textbf{.} 
In the main paper, we provide the impact and nonlocality measures in bar graphs for Tiny model only due to the limited space.
Here we provide the measures for all three different models of Tiny, Small, and Medium in Figs.~\ref{fig:impact_measure} and \ref{fig:nonlocality_measure}.

\clearpage

\smallbreak\noindent\textbf{Comparison with additional state-of-the-art baselines.} 
Table~\ref{table:sota_comparison_supp} summarizes model sizes, computational costs, and ImageNet performances of recent state-of-the-art methods and ours.
All the models are trained using images of $224 \times 224$ resolution.
PerViT with different model sizes achieves better (Tiny) or highly competitive (Small and Medium) performances compared with convnet and pyramidal ViT counterparts.
When compared with columnar vision transformers that use single-resolution feature maps, the proposed method performs the best.

\begin{table*}[h]
\begin{center}
\caption{Performance comparison on ImageNet~\cite{deng2009imagenet} with additional baselines.}
\scalebox{0.90}{
    \begin{tabular}{clccc}
    
    \toprule
    \multicolumn{2}{c}{Model} & Size (M) & FLOPs (G) & Top-1 (\%) \\
                
    \midrule\multirow{10}{*}{\shortstack{Fully\\Convolutional\\Networks}} 
    & ResNet-18~\cite{he2016deep}                    & 12  & 1.8 & 69.8 \\
    & RSB-ResNet-18~\cite{wightman2021rsbresnet}     & 12  & 1.8 & 70.6 \\
    \cline{2-5} \\[-2.0ex]
    
    & ResNet-50~\cite{he2016deep}                    & 25  & 4.1 & 78.5 \\ 
    & RSB-ResNet-34~\cite{wightman2021rsbresnet}     & 22  & 3.7 & 75.5 \\
    & RSB-ResNet-50~\cite{wightman2021rsbresnet}     & 26  & 4.1 & 79.8 \\
    & ConvNext-T~\cite{liu2022convnext}              & 29  & 4.5 & 82.1 \\
    \cline{2-5} \\[-2.0ex]
    
    & ResNet-101~\cite{he2016deep}                   & 45  & 7.9 & 79.8 \\ 
    & RSB-ResNet-101~\cite{wightman2021rsbresnet}    & 45  & 7.9 & 81.3 \\
    & RSB-ResNet-152~\cite{wightman2021rsbresnet}    & 60  & 12  & 81.8 \\
    & ConvNext-S~\cite{liu2022convnext}              & 50  & 8.7 & 83.1 \\
    
    \midrule\multirow{6}{*}{\shortstack{Spatial\\MLP-Mixer}} 
    & ResMLP-S12~\cite{touvron2021resmlp}            & 15  & 3.0 & 76.6 \\
    \cline{2-5} \\[-2.0ex]
    & gMLP-S~\cite{liu2021mlp}                       & 20  & 4.5 & 79.6 \\
    & ResMLP-S24~\cite{touvron2021resmlp}            & 30  & 6.0 & 79.4 \\
    \cline{2-5} \\[-2.0ex]
    & MLP-Mixer-B/16~\cite{tolstikhin2021mlpmixer}   & 59  & 13  & 76.4 \\
    & ResMLP-B24~\cite{touvron2021resmlp}            & 116 & 23  & 81.0 \\
    & gMLP-B~\cite{liu2021mlp}                       & 73  & 16  & 81.6 \\

    \midrule\multirow{17}{*}{\shortstack{Pyramidal\\Vision\\Transformers\\ \ \\ \ \\({\em multi-resolution})}} 
    & PVT-T~\cite{wang2021pvt}                       & 13  & 1.9 & 75.1 \\
    & CoaT-Lite-T~\cite{xu2021coat}                  & 5.7 & 1.6 & 77.5 \\
    & PoolFormer-S12~\cite{yu2021metaformer}         & 12  & 2.0 & 77.2 \\
    \cline{2-5} \\[-2.0ex]
    
    & CvT-13~\cite{wu2021cvt}                        & 20  & 4.5 & 81.6 \\
    & PVT-S~\cite{wang2021pvt}                       & 25  & 3.8 & 79.8 \\
    & Swin-T~\cite{liu2021swin}                      & 28  & 4.5 & 81.3 \\
    & CoaT-Lite-S~\cite{xu2021coat}                  & 20  & 4.0 & 81.9 \\
    & Focal-T~\cite{yang2021focal}                   & 29  & 4.9 & 82.2 \\
    & PoolFormer-S24~\cite{yu2021metaformer}         & 21  & 3.6 & 80.3 \\
    & PoolFormer-S36~\cite{yu2021metaformer}         & 31  & 5.2 & 81.4 \\
    \cline{2-5} \\[-2.0ex]
    
    & CvT-32~\cite{wu2021cvt}                        & 32  & 7.1 & 82.5 \\
    & PVT-M~\cite{wang2021pvt}                       & 44  & 6.7 & 81.2 \\
    & CoAtNet-1~\cite{dai2021coatnet}                & 42  & 8.4 & 83.3 \\
    & Swin-S~\cite{liu2021swin}                      & 50  & 8.7 & 83.0 \\
    & Focal-S~\cite{yang2021focal}                   & 51  & 9.1 & 83.5 \\
    & CoaT-Lite-M~\cite{xu2021coat}                  & 45  & 9.8 & 83.6 \\
    & PoolFormer-M36~\cite{yu2021metaformer}         & 56  & 9.1 & 82.1 \\
    & PoolFormer-M48~\cite{yu2021metaformer}         & 74  & 12  & 82.5 \\
    
    \midrule\multirow{16}{*}{\shortstack{Columnar\\Vision\\Transformers\\ \ \\ \ \\({\em single-resolution})}} 
    & DeiT-T~\cite{touvron2021deit}                 & 5.7 & 1.3 & 72.2 \\
    & ConViT-T~\cite{d2021convit}                   & 5.6 & 1.2 & 73.1 \\
    & ViT$_{C}$-1GF~\cite{xiao2021early}            & 4.6 & 1.1 & 75.3 \\
    & XCiT-T12/16~\cite{el2021xcit}                 & 7.0 & 1.2 & 77.1 \\
    & \textbf{PerViT-T (ours)}                      & 7.6 & 1.6 & 78.8 \\
    \cline{2-5} \\[-2.0ex]
    
    & DeiT-S~\cite{touvron2021deit}                 & 22  & 4.6 & 79.8 \\
    & ConViT-S~\cite{d2021convit}                   & 22  & 5.4 & 81.3 \\
    & ViT$_{C}$-4GF~\cite{xiao2021early}            & 18  & 4.0 & 81.4 \\
    & T2T-ViT$_{t}$-14~\cite{yuan2021t2tvit}        & 22  & 6.1 & 81.7 \\
    & XCiT-S12/16~\cite{el2021xcit}                 & 26  & 4.8 & 82.0 \\
    & \textbf{PerViT-S (ours)}                      & 21  & 4.4 & 82.1 \\
    \cline{2-5} \\[-2.0ex]
    
    & DeiT-B~\cite{touvron2021deit}                 & 86  & 18  & 81.8 \\
    & ConViT-S+~\cite{d2021convit}                  & 48  & 10  & 82.2 \\
    & T2T-ViT$_{t}$-24~\cite{yuan2021t2tvit}        & 64  & 15  & 82.6 \\
    & XCiT-S24/16~\cite{el2021xcit}                 & 48  & 9.1 & 82.6 \\
    & \textbf{PerViT-M (ours)}                      & 44  & 9.0 & 82.9 \\
        
    \bottomrule
    
    \end{tabular}
}
\vspace{0.0mm}
\label{table:sota_comparison_supp}
\vspace{0.0mm}
\end{center}
\end{table*}

\begin{table}
	\begin{minipage}{0.3\linewidth}
		\centering
        \caption{Ablation on $\Phi_{\text{p}}$ under different initializations.}
		\vspace{2.0mm}
        \scalebox{0.68}{
            \begin{tabular}{lccc}
            
            \toprule
            \multirow{2}{*}{\shortstack{Param.\\of $\Phi_\text{p}$}} & \multirow{2}{*}{\shortstack{Peripheral\\(ours)}} &         \multirow{2}{*}{Conv} & \multirow{2}{*}{Rand} \\
            & & & \\
            
            \midrule
            
            Trained & \textbf{78.8} & 78.6 & 78.5 \\
            Fixed   & \textbf{77.8} & 77.1 & 75.8 \\
            \midrule
            Absent  & \multicolumn{3}{c}{77.3} \\
                
            \bottomrule
            
            \end{tabular}
        }
        \vspace{0.0mm}
        \label{table:ablation_phip}
	\end{minipage}\hfill
	\begin{minipage}{0.66\linewidth}
		\centering
		\caption{Transfer learning results on CIFAR-10, CIFAR-100, and iNaturalist-19.}
		\vspace{3.0mm}
        \scalebox{0.71}{
		\begin{tabular}{lcccccc}
			\toprule
			Model           & Size (M) & FLOPs (G) & CIFAR-10 & CIFAR-100 & iNAT-19 \\
			\midrule
			ViT-L~\cite{dosovitskiy2021vit}           & 307 & 117 & 97.9 & 86.4 & -  \\
			DeiT-B~\cite{touvron2021deit}          & 86 & 18 & \textbf{99.1} & 90.8  & 77.7  \\
			\midrule
			PerViT-M (ours) & \textbf{44} & \textbf{9} & \textbf{99.1} & \textbf{91.4} & \textbf{78.5} \\
			
			\bottomrule
		\end{tabular}
		}
		\vspace{2.0mm}
		\label{table:transfer_learning_results}
	\end{minipage}
	\vspace{-5.0mm}
\end{table}

\smallbreak\noindent\textbf{Additional ablation on} $\Phi_{\text{p}}$. 
To highlight the benefits of learning $\Phi_{\text{p}}$, we conduct experiments using PerViT-T with parameters of $\Phi_{\text{p}}$ fixed during training under three different initialization methods of peripheral, conv, and rand.
Table~\ref{table:ablation_phip} summarizes the results.
Fixing $\Phi_{\text{p}}$ parameters damages performance for all three intializations, verifying the efficacy of learning diverse position-based attentions across different layers and heads.
We observe that conv and rand inits perform poorly compared to PerViT-T without $\Phi_{\text{p}}$, \eg, model (b) in Tab.~\ref{table:ablation_main} (77.3\%);
we suspect that fixed $\Phi_{\text{p}}$ with conv and rand inits only provide local and noisy attentions respectively while PerViT without $\Phi_{\text{p}}$ has no such strong restrictions.

\smallbreak\noindent\textbf{Transfer learning results.} 
We verify the robustness of the proposed method by comparing the PerViT-M with baseline models~\cite{dosovitskiy2021vit, touvron2021deit} on different transfer learning task with ImageNet pre-training in Tab.~\ref{table:transfer_learning_results}.
We finetune trained PerViT-M on CIFAR-10, CIFAR-100, and iNaturalist-19, following the same training recipes of DeiT~\cite{touvron2021deit}.
Even with significantly lower complexity than~\cite{dosovitskiy2021vit, touvron2021deit}, our method surpasses baselines by approximately 1\%p on CIFAR-100 and iNaturalist19 while performing on par with \cite{touvron2021deit} on CIFAR-10.

\section{Model Layout Details}
\label{sec:supp_layout_details}
This paper explores PerViT with three different model sizes: Tiny, Small, and Medium.
Each model consists total 12 blocks of layers ($N_l = 12$) divided into 4 stages each of which uses different channel sizes ($D_{\text{emb}} = D_hN_h$).
The expansion ratio of the MLP layer is set to 4 following DeiT~\cite{touvron2021deit}.
We use normalized coordinates to ensure numerical stability during training such that $[-1, -1]^{\top} \leq \mathbf{q}, \mathbf{k} \leq [1, 1]^{\top}$.
The input and hidden dimensions of peripheral projections are set as $D_{\text{r}} = D_\text{hid} = 4N_h$. 
We set $K=3$ to capture neighboring distance representations ($\mathcal{N}$) in the peripheral projections;
in experiments, increasing the neighborhood size ($K > 3$) hardly brought improvements.
We hypothesize the $3 \times 3$ window is sufficient for the model to capture the spatial dimension of the input images.
Similarly to our baseline, \eg, DeiT~\cite{touvron2021deit}, the largest model (PerViT-M) require a few days of training on 8 V100 GPUs.

Table~\ref{table:model_layout} summarizes architecture details for the three models.
As seen in the last two columns, despite the small size of the proposed peripheral position encoding $\Phi_{\text{p}}$ ($\leq$0.6\% of the network size), it significantly boosts top-1 ImageNet accuracy (1.4\%p$\sim$4.2\%p improvements with $\Phi_{\text{p}}$ for Tiny) as shown in Tab.~\ref{table:ablation_main} of the main paper, verifying its effectiveness in terms of both accuracy and efficiency.

\begin{table*}[h]
\begin{center}
\vspace{-1.0mm}
\caption{Model layout for PerViT Tiny (T), Small (S), and Medium (M) models.}
\vspace{-1.0mm}
\scalebox{0.74}{
    \begin{tabular}{lcccccccc}
    
    \toprule
    \multirow{2}{*}{\shortstack{PerViT\\model}} & \multirow{2}{*}{\shortstack{\# heads\\($N_h$)}} & \multirow{2}{*}{\shortstack{\# layers\\at each stage}} & \multirow{2}{*}{\shortstack{Channel sizes\\at each stage ($D_{\text{emb}}$)}} & \multirow{2}{*}{\shortstack{Channel sizes in\\conv. patch embedding}} & \multirow{2}{*}{\shortstack{Size\\(M)}} & \multirow{2}{*}{\shortstack{FLOPs\\(G)}} & \multicolumn{2}{c}{Attention $\Phi_{\text{p}}$} \\
    & & & & & & & Size (M) & Ratio \\
    
    \midrule
    Tiny          & 4  & [2, 2, 6, 2] & [128, 192, 224, 280] & [48, 64, 96, 128] & 7.6 & 1.6 & 0.04M & 0.3\% \\
    Small         & 8  & [2, 2, 6, 2] & [272, 320, 368, 464] & [64, 128, 192, 262] & 21.3 & 4.4 & 0.14M & 0.6\% \\
    Medium        & 12 & [2, 2, 6, 2] & [312, 468, 540, 684] & [64, 192, 256, 312] & 43.7 & 9.0 & 0.31M & 0.6\% \\

    \bottomrule
    
    \end{tabular}
}
\vspace{-1.0mm}
\label{table:model_layout}
\vspace{-3.0mm}
\end{center}
\end{table*}

\clearpage

\section{Training Details}
\label{sec:supp_training_details}
We follow similar training recipes of DeiT~\cite{touvron2021deit}, a hyperparameter-optimized version of ViT~\cite{dosovitskiy2021vit}.
The stochastic depth rate is used in Small and Medium models where we set them to 0.1 and 0.2 respectively.
For Medium model, we use 20 warmup epochs as smaller warmup epoch did not converge in our experiments.
Table~\ref{table:trainig_detail} summarizes the details on our training recipe.

\begin{table*}[h]
\begin{center}
\vspace{-1.0mm}
\caption{Training parameters for PerViT Tiny (T), Small (S), and Medium (M) and DeiT-B~\cite{touvron2021deit}.}
\vspace{-0.0mm}
\scalebox{0.95}{
    \begin{tabular}{lrclc}
    
    \toprule
    Methods & \multicolumn{3}{c}{PerViT} & DeiT-B~\cite{touvron2021deit} \\
    
    \midrule
    Epochs & \multicolumn{3}{c}{300} & 300 \\
    \midrule
    Batch size              & \multicolumn{3}{c}{1024} & 1024 \\
    Optimizer               & \multicolumn{3}{c}{AdamW} & AdamW\\
    Learning rate           & \multicolumn{3}{c}{$0.0005 \times \frac{batch size}{512}$} & $0.0005 \times \frac{batch size}{512}$ \\
    Learning rate decay     & \multicolumn{3}{c}{cosine} & cosine\\
    Weight decay            & 0.03 (T) & 0.05 (S) & 0.05 (M) & 0.05 \\
    Warmup epochs           & 5 (T)&5 (S)& 20 (M) & 5 \\
    \midrule
    Label smoothing $\epsilon$   & \multicolumn{3}{c}{0.1} & 0.1 \\
    Dropout                      & \multicolumn{3}{c}{\xmarkb} & \xmarkb \\
    Stoch. Depth                 & 0.0 (T)& 0.1 (S)& 0.2 (M) & 0.1 \\
    Repeated Aug                 & \multicolumn{3}{c}{\cmarkb} & \cmarkb \\
    Gradient clip                & \multicolumn{3}{c}{\xmarkb} & \xmarkb \\
    \midrule
    Rand Augment                 & \multicolumn{3}{c}{9/0.5} & 9/0.5 \\
    Mixup prob.                  & \multicolumn{3}{c}{0.8} & 0.8 \\
    Cutmix prob.                 & \multicolumn{3}{c}{1.0} & 1.0 \\
    Erasing prob.                & \multicolumn{3}{c}{0.25} & 0.25 \\

    \bottomrule
    
    \end{tabular}
}
\vspace{-1.0mm}
\label{table:trainig_detail}
\vspace{-0.0mm}
\end{center}
\end{table*}


\section{Societal Implications and Broader Impacts}
\label{sec:supp_broader_impacts}

The focus of this work is model exploration \& development for image classification task, providing an original direction towards combining human peripheral vision with transformer-based architecture.
To the best of our knowledge, this work poses no immediate negative impact on society other than environmental concerns related to CO$_2$ emisssion;
training vision transformers on large-scale dataset~\cite{deng2009imagenet} from scratch demands a substantial amount of computational resources, \eg, GPUs.

Our work may inspire biologically-inspired computer vision researches, which would potentially promote the creation of stronger machine vision.
While we have focused on image classification task, we believe that the idea can be broadly applicable for high-level vision applications such as object detection \& segmentation, and action recognition.
We leave this to future work.

{\small
    \bibliographystyle{abbrv}
    \bibliography{egbib}
}


\end{document}